\documentclass[10pt,twocolumn,letterpaper]{article}

\usepackage[pagenumbers]{wacv} %

\definecolor{wacvblue}{rgb}{0.21,0.49,0.74}
\usepackage[pagebackref,breaklinks,colorlinks,allcolors=wacvblue]{hyperref}

\usepackage{microtype}
\usepackage{graphicx}
\usepackage{booktabs} %

\usepackage{amssymb}
\usepackage{mathtools}
\usepackage{amsthm}
\usepackage[acronym]{glossaries}
\newacronym{hv}{HiVec}{High Vectorization}

\usepackage[capitalize,noabbrev]{cleveref}

\theoremstyle{plain}

\theoremstyle{definition}

\theoremstyle{remark}

\usepackage{amsmath,amsfonts,bm}

\def\eqref#1{equation~\ref{#1}}

\def\1{\bm{1}}

\DeclareMathAlphabet{\mathsfit}{\encodingdefault}{\sfdefault}{m}{sl}
\SetMathAlphabet{\mathsfit}{bold}{\encodingdefault}{\sfdefault}{bx}{n}

\usepackage{graphicx}
\usepackage{amsmath}
\usepackage{amssymb}
\usepackage{booktabs}

\usepackage{url}

\usepackage{graphicx, color}
\usepackage{tikz}
\usepackage{comment}
\usepackage{amsmath,amssymb} %
\usepackage{color}
\usepackage{ragged2e}
\usepackage{xcolor}         %
\usepackage{graphicx}
\usepackage{multirow}

\usepackage{subcaption}

\usepackage{caption}
\usepackage{float}

\usepackage{url}            %
\usepackage{booktabs}       %
\usepackage{amsmath}
\usepackage{amsfonts}       %

\usepackage{nicefrac}       %
\usepackage{microtype}      %
\usepackage{nicefrac}       %
\usepackage{wrapfig}

\usepackage{multirow}
\usepackage{makecell}
\usepackage{threeparttable}
\usepackage{sidecap}
\usepackage{caption}

\newcommand{\x}{\mathbf{x}}

\newcommand{\y}{\mathbf{y}}

\newcommand{\btheta}{\boldsymbol{\theta}}
\newcommand{\bphi}{\boldsymbol{\phi}}

\usepackage{colortbl} %
\definecolor{mColor1}{rgb}{0.9,0.9,0.9}
\definecolor{mColor2}{rgb}{0.95,0.95,0.95}
\definecolor{non-photoblue}{rgb}{0.64, 0.87, 0.93}
\definecolor{lightblue}{rgb}{0.81, 0.94, 1.0}
\usepackage{tikz}
\usepackage{pifont}

\definecolor{mColor1}{rgb}{0.9,0.9,0.9}
\definecolor{mColor2}{rgb}{0.95,0.95,0.95}
\definecolor{non-photoblue}{rgb}{0.64, 0.87, 0.93}
\definecolor{lightblue}{rgb}{0.81, 0.94, 1.0}
\definecolor{lightorange}{rgb}{0.965, 0.835, 0.71}

\definecolor{mygreen}{rgb}{0.000, 0.392, 0.000}

\usepackage{orcidlink}
\definecolor{mypurple}{rgb}{0.502, 0.000, 0.502}

\newcommand{\blue}[1]{\textcolor{black}{#1}}
\newcommand{\bluet}[1]{\textcolor{black}{#1}}

\newcommand{\bluett}[1]{\textcolor{black}{#1}}

\newcommand{\purple}[1]{\textcolor{black}{#1}}

\newcommand{\ple}[1]{\textcolor{black}{#1}}

\newcommand{\pinktt}[1]{\textcolor{black}{#1}}

\usepackage{algorithm}
\usepackage{algorithmic}
\usepackage{pifont}

\usepackage[capitalize]{cleveref}
\crefname{section}{Sec.}{Secs.}
\Crefname{section}{Section}{Sections}
\Crefname{table}{Table}{Tables}
\crefname{table}{Tab.}{Tabs.}

\usepackage{booktabs} %
\usepackage{multirow} %
\usepackage{graphicx} %

\usepackage{soul}
\usepackage{xcolor}

\sethlcolor{yellow} %

\definecolor{cgreen}{rgb}{0.2,0.6,1}

\definecolor{darkgreen}{RGB}{0,100,0}

\newcommand{\inc}[1]{%
  #1 \rlap{\raisebox{0.1ex}{\textcolor{darkgreen}{$\blacktriangle$}}}%
}

\newcommand{\incsymb}{\rlap{\raisebox{0.1ex}{\textcolor{darkgreen}{$\blacktriangle$}}}\hspace{0.2em}}

\usepackage{etoc}

\def\wacvPaperID{094} %
\def\confName{WACV}
\def\confYear{2026}

\title{Hierarchical Adaptive networks with Task vectors  \\for Test-Time Adaptation}

\author{
  Sameer Ambekar\textsuperscript{1,2,4,5},
  Marta Hasny\textsuperscript{1,2},
  Laura Daza\textsuperscript{1,2},
  Daniel M. Lang\textsuperscript{1,2}\thanks{Shared last-authorship},  %
  Julia A. Schnabel\textsuperscript{1,2,3,4,5}\footnotemark[\value{footnote}]\\  %
  \textsuperscript{1}School of Computation, Information and Technology, Technical University of Munich, Germany \\
  \textsuperscript{2}Institute of Machine Learning in Biomedical Imaging, Helmholtz Munich, Germany \\
  \textsuperscript{3}School of Biomedical Engineering and Imaging Sciences, King's College London, UK \\
  \textsuperscript{4}Munich Center for Machine Learning (MCML)\\
  \textsuperscript{5}relAI – Konrad Zuse School of Excellence in Reliable AI\\
}

\begin{document}

\newpage
\clearpage
\maketitle

\begin{abstract}

Test-time adaptation allows pretrained models to adjust to incoming data streams, addressing distribution shifts between source and target domains. However, standard methods rely on single-dimensional linear classification layers, which often fail to handle diverse and complex shifts. We propose \ple{Hierarchical Adaptive Networks with Task Vectors (Hi-Vec), which leverages multiple layers of increasing size} for dynamic test-time adaptation. By decomposing the encoder's representation space into such hierarchically organized layers, \ple{Hi-Vec, in a plug-and-play manner, allows existing methods to adapt to shifts of varying complexity.} Our contributions are threefold: \bluet{First}, we propose dynamic layer selection for automatic identification of the optimal layer for adaptation to each test batch. \bluet{Second}, we propose a mechanism that merges weights from the dynamic layer to other layers, \ple{ensuring all layers receive target information.} \bluet{Third}, \ple{we propose linear layer agreement that acts as a gating function, preventing erroneous fine-tuning by adaptation on noisy batches}. We rigorously evaluate the performance of Hi-Vec in challenging scenarios and on multiple target datasets, proving its strong capability to advance state-of-the-art methods. \ple{Our results show that Hi-Vec improves robustness, addresses uncertainty, and handles limited batch sizes and increased outlier rates. Code: \url{https://github.com/ambekarsameer96/Hi-Vec}}

\end{abstract}

\section{Introduction}
The test-time adaptation paradigm was introduced to mitigate performance degradation of deep learning models when faced with new data that differ from the training distribution~\cite{wang2021tent,iwasawa2021test,liu2021ttt++,weijler2024ttt, liang2020we}. These techniques adjusts the model's parameters or internal representations on test data, aiming to handle the distribution gap between training and inference, and to manage previously unseen shifts. Recent advances show that test-time adaptation can effectively tackle challenges such as unexpected image corruptions~\cite{wang2021tent,liang2020we,Cho_2025_WACV} and domain changes~\cite{xiao2020vaebm,iwasawa2021test,Bahri_2025_WACV, ambekarvariational}, in tasks like classification~\cite{liu2021ttt++,lee2024entropy} and semantic segmentation~\cite{bateson2022test,varsavsky2020test}.

\begin{figure*}[ht!]
\centering
\includegraphics[width=1.0\linewidth, height=1.0\textheight, keepaspectratio]{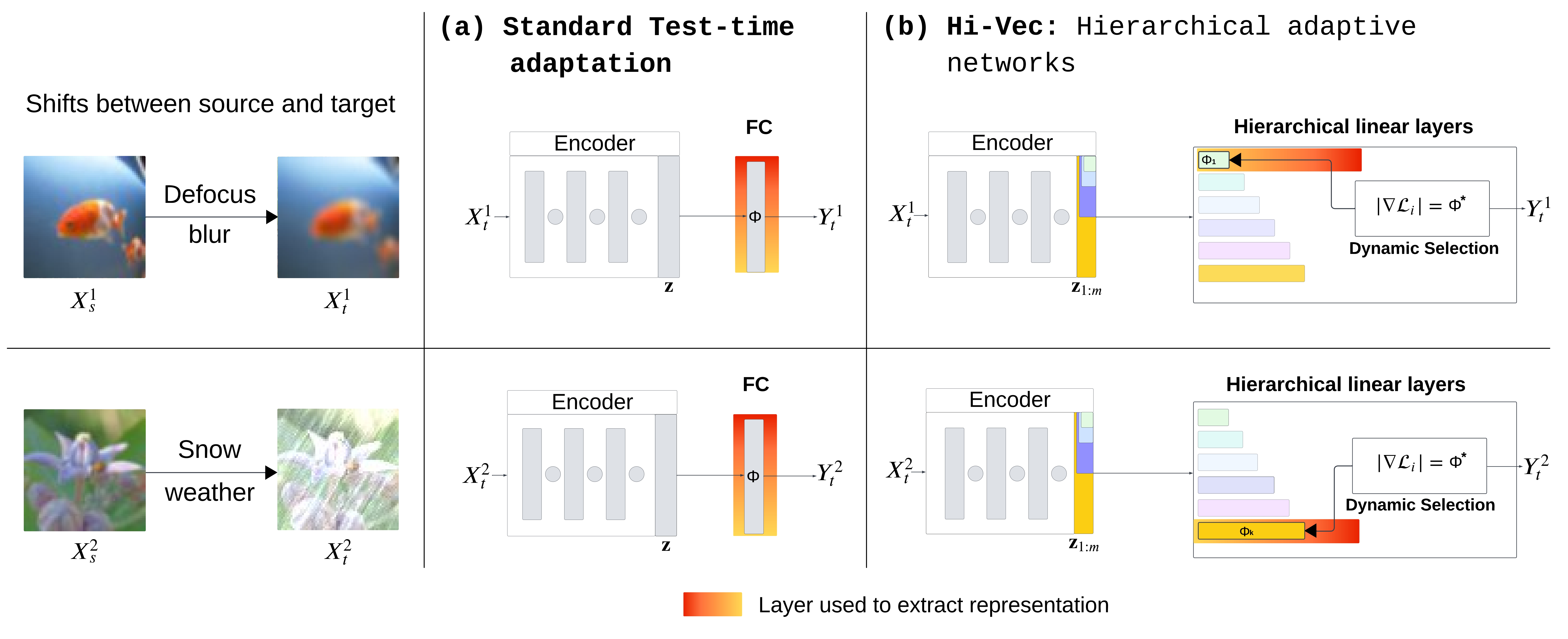}

\caption{\textbf{Standard methods and Hi-Vec alongside examples of shifts. } (a) Standard test-time adaptation methods adjust the same set of parameters and rely on the representation of a single-dimensional linear layer, which is forced to handle all kinds of domain shifts at the same time.
(b) Hi-Vec introduces hierarchical linear layers featuring coarse-to-fine representations, and uses dynamic selection, which allows for individual handling by identification of the layer most suitable to address a specific domain shift. (detailed architecture in Figure~\ref{fig2:arch}).}
\label{fig1:all}
\vspace{-4mm}
\end{figure*}

A major challenge in test-time adaptation originates from the limited flexibility of representations learned by standard deep learning methods, which typically use a single linear layer after the encoder to project these representations into the output space, as shown in Fig.~\ref{fig1:all}a. This setup forces all features through a single projection, leading to \textit{information diffusion}~\cite{soudry2018implicit}, where fine-grained details become blurred across layers. Additionally, the final layer of the encoder may experience rank collapse, a phenomenon in which representations lose expressivity and become overly constrained~\citep{feng2022rank,kusupati2022matryoshka}. {In response to the aforementioned drawbacks, \citet{kusupati2022matryoshka} propose Matryoshka Representation Learning (MRL), which attaches multiple linear layers of different sizes to the encoder. Each layer in MRL maps a progressively larger subset of the encoder’s features to the output space. This hierarchical design promotes coarse-to-fine feature extraction, enhancing the overall expressivity and flexibility of the learned representation.} 

Building on MRL, we introduce \textbf{Hi-Vec} (Hierarchical Layers with Task Vectors), a novel framework designed to handle diverse distribution shifts during test time. Hi-Vec leverages hierarchical representations to dynamically identify the optimal level of granularity required to handle the current distribution shift (see also Fig.~\ref{fig1:all}b).  This forms a versatile backbone for test-time adaptation, accommodating a broad range of data distributions. \noindent {Our key contributions are as follows:}

\begin{itemize}
    \item Dynamic layer selection, which automatically identifies the most suitable layer for adapting to each test batch. This addresses the problem of uniform parameter adjustments regardless of shift complexity, ensuring targeted adaptation for every batch. 
    \item Target information sharing via task vectors, which merges adapted weights from the selected layer to other linear layers for cross-layer coherence. This facilitates efficient information dissemination between linear layers.
    \item Hierarchical Layer Agreement, a gating strategy that moderates per-batch adaptation to avoid fine-tuning on noisy samples. This is crucial for practical test-time adaptation, as it prevents erroneous model updates and enhances overall robustness to noisy test data. \\
\end{itemize}

In Section~\ref{Section:Preliminary}, we introduce the problem setup and hierarchical linear layer configuration. Section~\ref{section:layers_at_test_time} presents our method's theoretical foundation and its three main components. Section~\ref{sec:exps} details experiments integrating Hi-Vec with four state-of-the-art test-time adaptation methods. We evaluate two challenging scenarios: (i) Section~\ref{Section:Outlier} assesses outlier-aware settings, where test batches mix target samples with out-of-distribution ones from unrelated datasets (ii) Section~\ref{Section:Spurious} examines spurious correlations, using target datasets differing from the source domain, forcing adaptation without misleading on source features. In Section~\ref{Section:benefits}, we detail the benefits and insights of using our framework.

\section{Related Work}

\textbf{Domain Adaptation}  
Distribution shifts between training and testing domains often degrade model performance, necessitating schemes like domain adaptation and domain generalization. Domain adaptation assumes access to target data during source training and aligns distributions using techniques such as adversarial training or feature alignment~\cite{pandey2020unsupervised, vu2019advent, hoffman2018cycada, ben2006analysis, yang2024generalized}. However, this approach is impractical without prior target data access. Domain generalization counters this by learning domain-invariant features sans target data during training~\cite{pandey2021generalization, gulrajani2020search, muandet2013domain}. While effective in certain scenarios, these methods struggle with test data arriving incrementally in small batches. Source-free domain adaptation extends them by adjusting models to target data at test time without source access~\cite{liang2020we, kundu2022balancing, wang2022exploring}. Taking it further, test-time adaptation refines the model at each inference step.

\noindent \textbf{Test-Time Adaptation.}
Test-time adaptation allows models to address distribution shifts by adapting and predicting simultaneously on incoming streams of unlabeled target data~\cite{chen2022ost,choi2021test,min2023meta,xiao2022learning,sahoo2024layer,xiao2024beyond,liang2023comprehensive,xiao2024any,xiao2023energy,boudiaf2022parameter,kim2025testtime,vray2025reservoirtta,ambekar2025gf,Yang_2024_CVPR,lei2025ttvd,du2024unitta}. Unlike domain adaptation~\cite{pandey2020unsupervised,vu2019advent,hoffman2018cycada} and domain generalization~\cite{muandet2013domain,gulrajani2020search}, test-time adaptation uniquely operates during inference, making it suitable for real-world scenarios where target data is unavailable during training and arrives sequentially. Standard fine-tuning approaches typically adjust a fixed set of parameters, such as batch normalization statistics~\cite{zhang2021adaptive,nado2020evaluating}, full model parameters~\cite{wang2021tent,liang2020we}, or single-dimensional linear layers~\cite{jang2022test,iwasawa2021test,zhang2023adanpc}. \ple{Beyond these, recent works have explored overall layer selection~\cite{sahoo2024layer} and surgical fine-tuning~\cite{lee2022surgical} for targeted adaptation.} These methods have been evaluated on unseen scenarios, including noise-corrupted datasets~\cite{wang2021tent,goyal2022test} and datasets with varying domain information~\cite{iwasawa2021test,liang2020we}. \purple{Recently, \citet{yu2024stamp} incorporated outliers into the evaluation protocol and introduced a memory-based approach with self-weighted entropy to handle diverse distribution shifts. However, like standard test-time adaptation methods, their approach relies on a single-dimensional linear layer and repeatedly uses the same batch norm parameters, limiting feature expressivity.}

\noindent \bluet{Moreover, beyond fully test-time adaptation methods~\cite{xiao2024beyond}, several approaches propose modifying the training process by learning a surrogate task~\cite{liu2021ttt++,xiao2023energy} or new parameters~\cite{lim2023ttn,zhou2022learning}. In this work, we also propose to modify the training process by learning multiple-sized linear layers and allowing existing fully test-time adaptation methods~\cite{wang2021tent,yu2024stamp,niu2022towards,lee2024entropy} to adapt to diverse shifts.}

\noindent \textbf{Model Merging.} Pretrained models are widely fine-tuned for downstream tasks to mitigate bias, align with specific objectives, or integrate additional information across supervised and unsupervised settings. Recent advances have established model merging~\cite{Survey_ModelAggregationinFL_2023,surgery_icml2024,frankle2020linear,yang2024model} as a powerful technique for combining knowledge from multiple pretrained models without requiring access to source data or extensive computations. For instance, \citet{ilharco2022editing} introduced task vectors, which use arithmetic operations on the weights of models with the same architectures to allow tasks like knowledge combination, forgetting, and improved domain generalization. Similarly, \citet{stoica2023zipit} explored merging the models trained for different problem settings by identifying and combining shared features through targeted weight manipulation. Other methods, such as those in~\cite{Modelsoups_ICML2022,SparseModelSoups,adaptersoup_EACL2023,rame2023modelratatouille}, average model weights to produce an enhanced, unified model. Furthermore, \citet{daheim2023model} demonstrated the fusion of networks trained on distinct datasets by resolving gradient mismatches. In contrast, while previous works typically merge models of uniform dimensions, our framework uniquely merges layers of different dimensionality for test-time adaptation to diverse shifts.

\section{Background}
\label{Section:Preliminary}
\subsection{Notations and Baseline methods} 
Test-time adaptation seeks to adapt a model $\btheta_{s}$ trained on a source domain $D_s$ to perform effectively on an unseen target domain $D_t$. While $D_s$ consists of data-label pairs $(\mathbf{x}_s, y_s)$ drawn from a uniform distribution, $D_t$ may exhibit diverse shifts relative to $D_s$. In standard test-time adaptation settings, $D_t$ is assumed to follow a single distribution. However, realistic scenarios often introduce complex challenges, such as disruptive or noisy components within $D_t$. \emph{Outlier shifts} emerge when test batches contain a small number of samples from an alternative distribution $D_t^{(o)}$, significantly different from the source. \emph{Spurious correlation shifts} occur when samples in $D_t^{(s)}$ display non-causal associations, such as background features tied non-causally to classes.

\begin{figure*}[ht!]
\centering
\includegraphics[width=1.01\linewidth, height=1.0\textheight, keepaspectratio]{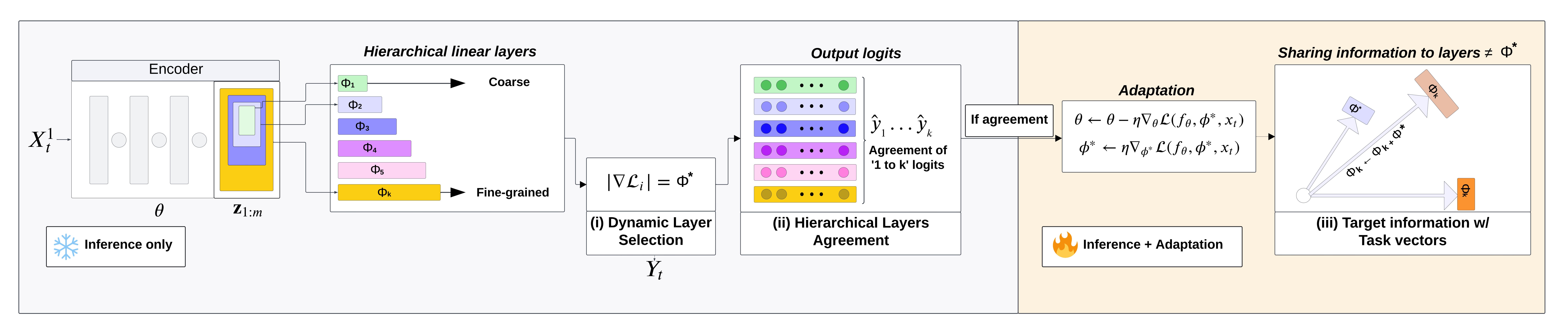}

\caption{\textbf{Illustration of Hi-Vec.} Our framework introduces (i) \textit{Dynamic selection} of hierarchical linear layers to find the optimal linear layer for the specific type of distribution shift in each batch. Next, through (ii) \textit{Hierarchical layer agreement}, it evaluates logit consistency across multiple representations to decide if adaptation is needed. Finally, if agreement, (iii) \textit{Target information sharing via task vectors} and adaptation are performed; otherwise, it proceeds directly to inference, minimizing computational overhead and erroneous finetuning.}
\label{fig2:arch}
\vspace{-4mm}
\end{figure*}

\blue{The source model $\btheta_{s}$ is initially trained by minimizing the empirical risk (loss), i.e., $\min_{\theta} \mathbb{E}_{x_{s} \sim \mathcal{D}_{s}} \big[ \mathcal{L}_{s}(f_{\theta}(x_s)) \big]$, with $\mathcal{L}_{s}$ typically representing cross-entropy~\cite{gulrajani2020search} or a surrogate loss~\cite{vu2019advent,shu2018dirt,hoffman2018cycada}. To handle distribution shifts at test time, these methods fine-tune specific components of the model, such as batch normalization layers ($\beta, \gamma$)~\cite{wang2021tent,goyal2022test}, linear heads ($\bphi$)~\cite{wang2021tent,iwasawa2021test}, or the full set of parameters ($\btheta$)~\cite{liu2021ttt++}. This is achieved by minimizing an unsupervised loss $\mathcal{L}_t$ based on entropy~\cite{wang2021tent,zhang2021memo,zhou2021bayesian,goyal2022test}, self-supervision~\cite{xiao2022learning}, or surrogate tasks~\cite{liu2021ttt++,li2018metalearning}, yielding the optimization objective:
\begin{equation}
\min_{\{\beta, \gamma, \btheta, \bphi\}} 
\mathbb{E}_{x_{t} \sim \mathcal{D}_{t}} 
\big[ 
\mathcal{L}_{t}(f_{\{\beta, \gamma, \btheta, \bphi\}}(x_t))
\big]
\end{equation}}

\purple{However, as shown in Fig.~\ref{fig1:all}a, a single linear head ($\phi$) that maps all the learned features to the output space, restricting their flexibility and granularity. Furthermore, this approach risks diffusing critical information from the encoder~\cite{kusupati2022matryoshka}, which erodes the model's ability to learn across high- to low-dimensional scales.} Consequently, test-time adaptation methods that rely on these single mappings often fail to handle varying distribution shifts effectively~\cite{lee2022surgical}. This increases the risk of overfitting and restricts inter-batch adaptability. Moreover, repeatedly adapting the same parameters ($\gamma, \beta, \phi$) across all batches overlooks batch-specific complexities, resulting in suboptimal performance under diverse test-time conditions.

\vspace{-0.5mm}
\subsection{Hierarchical Layers}
\label{section:hierarchial_linear_layers}

\blue{To learn coarse-to-fine representations, we adopt Matryoshka layers~\cite{kusupati2022matryoshka}, which we refer to as hierarchical linear layers \bluett{due to their increasing dimensionality and representational design.} These layers decompose the encoder's latent representation $\mathbf{z} \in \mathbb{R}^d$ into a set of subspaces $\mathbf{z}_{\mathcal{M}} = \{\mathbf{z}_{1:k} \in \mathbb{R}^k \mid k \leq d \}$, organizing the feature space into increasingly refined levels. This hierarchical organization enhances adaptability and encourages multi-scale representation learning.} 

Formally, let $\mathbf{z}_{1:m} \in \mathbb{R}^m$ be the first $m$ dimensions of $\mathbf{z}$, where $m \in \mathcal{M}$ and $\mathcal{M} = \{m_1, m_2, \dots, m_k\}$ satisfies $m_1 < m_2 < \dots < m_k = d$. Furthermore, let $\Phi = \{\bphi_{1}, \bphi_{2}, \dots, \bphi_{k}\}$ be the set of hierarchical linear layers that map the subspaces directly to the output space. The logits for layer $\bphi_i$ are defined as $\mathbf{y}_{m_i} = W_{1:m_i}^\top \mathbf{z}_{1:m_i}$, where $W_{1:m_i} \in \mathbb{R}^{m_i \times C}$ are the weights for $C$ classes. During source training, the model optimizes an empirical risk minimization (ERM) objective~\cite{gulrajani2020search} by minimizing the cross-entropy loss ($\mathcal{L}_{\mathrm{CE}}$) across all hierarchical layers and the encoder:
\begin{equation}
\label{source_train}
\begin{aligned}
& \btheta, \{W_{\bphi}\}_{\bphi \in \Phi} = \\
& \quad \arg\min_{\btheta, \{W_{\bphi}\}_{\bphi \in \Phi}} \mathbb{E}_{(\mathbf{x}_s, \mathbf{y}_s) \sim \mathcal{D}_s}  
\left[
\sum_{m \in \mathcal{M}} \mathcal{L}_{\mathrm{CE}}(\mathbf{y}_{m_i}, \mathbf{y}_s)
\right]
\end{aligned}
\end{equation}

\section{Methodology: Hierarchical Layers for Test-Time Adaptation}
\label{section:layers_at_test_time}

{Hi-Vec leverages the hierarchical coarse-to-fine structure learned during source training to address diverse distribution shifts in the target domain $\mathcal{T} = {D_t}$, where different dimensional embeddings capture varying levels of abstraction~\cite{lindeberg2013scale,kusupati2022matryoshka}. Therefore, each hierarchical linear layer $\bphi_{1}, \bphi_{2}, \dots, \bphi_{k}$ encodes a distinct level of representational granularity, making it inherently suitable for adaptive feature learning\bluett{~\cite{kusupati2022matryoshka}}.}

{This hierarchical organization follows established principles in representation learning~\cite{feng2022rank,zhang2023adanpc,kusupati2022matryoshka}. Lower-dimensional layers naturally preserve the most essential and robust features due to their compressed nature, as information bottlenecks tend to filter out unwanted features and retain core semantic content~\cite{jain1999data,harlev2023exploring}. Higher-dimensional layers maintain increasingly detailed representations that can capture more nuanced patterns and variations~\cite{cordier2022test,iwasawa2021test}.}

{Hi-Vec addresses test-time adaptation by dynamically selecting the most appropriate representational granularity for each encountered distribution shift, leveraging the natural stability of lower-dimensional representations for robust adaptation while using higher-dimensional layers when complex domain gaps. This hierarchical approach enables the framework to automatically match the adaptation complexity to the shift complexity, providing both computational efficiency and adaptation stability across diverse target domains containing multiple types of shifts and noisy outliers.}

\subsection{Dynamic Layer Selection}
\label{section:gradient}

To identify the most suitable layer for each batch dynamically, we compute the gradient norm for each layer independently, using an \bluett{unsupervised} entropy minimization loss~\cite{wang2021tent}. The layer with the lowest gradient norm is selected for prediction and adaptation, as it indicates best alignment. Formally, let \(\|\nabla_{W_{\bphi}} \mathcal{L}\|\) denote the gradient norm with respect to the parameters of layer \(\bphi\). The optimal layer \(\bphi^*\) is selected as:
\begin{equation}
\label{eq:gradient}
\bphi^{*} = \arg\min_{\bphi \in \Phi} \|\nabla_{W_{\bphi}} \mathcal{L}\|
\end{equation}

\purple{Once \(\bphi^*\) is selected, the encoder \(\btheta\) and the chosen hierarchical layer \(\bphi^*\) can be fine-tuned using existing test-time adaptation methods~\cite{wang2021tent,lee2024entropy,yu2024stamp,niu2022efficient}. In particular, any test-time method relying on an unsupervised loss for parameter updates is seamlessly compatible with our framework.} As a plug-and-play extension, Hi-Vec enhances existing methods through hierarchical layers and dynamic selection promoting targeted adaptation per batch. \pinktt{We detail the rationale for choosing the lower gradient norm, since it identifies the layer requiring minimal adaptation in Supplementary A.4.}

However, while dynamic selection improves efficient adaptation, unselected layers (\(\bphi \neq \bphi^*\)) would normally remain unchanged, retaining outdated knowledge, misaligning with future target batches. To mitigate this, building on task vectors~\cite{ilharco2022editing}, we propose target information sharing mechanism to propagate target-specific updates across layers.

\subsection{Target Information sharing with Task Vectors}
\label{section:knowledge_sharing}
To allow for dissemination of target information from the selected layer \(\bphi^*\) to other layers, we propose a task vectors~\cite{ilharco2022editing} based mechanism that does not require gradient computation or backpropagation. Unlike uniform merging of prior works, our framework handles hierarchical layers of varying dimensions through dimension-aware updates, merging subsets of compatible weights. \purple{In this framework, each hierarchical layer \(\bphi\) acts as a task vector \(\mathbf{v}_{\bphi}\)}. We compute affinities using cosine similarity~\cite{HAN201239} and identify similar layers exceeding a threshold \(\tau\):
\begin{equation}
\mathcal{S} = \{\bphi \in \Phi \mid \text{Sim}(\mathbf{v}_{\bphi^*}, \mathbf{v}_{\bphi}) > \tau\}
\end{equation}
The weights of \(\bphi^*\), denoted \(W_{\bphi^*}\), are then propagated to layers in \(\mathcal{S}\) using:
\begin{equation}
\label{eq:update-rule}
W_{\bphi}[\bphi^*] \gets W_{\bphi^*} + \alpha \, W_{\bphi}[\bphi^*]
\end{equation}
where \(W_{\bphi}[\bphi^*]\) represents the subset of weights in \(W_{\bphi}\) corresponding to those in \(W_{\bphi^*}\), and \(\alpha\) is a scaling factor. All other weights in \(W_{\bphi}\) remain unchanged. \pinktt{A detailed formulation of the notation introduced here can be found in Supplementary A.9.} This direct propagation of updates from \(\bphi^*\) to similar layers is theoretically supported by the principles of linear mode connectivity~\cite{frankle2020linear}, because the layers are trained jointly on source data, creating smooth, connected paths through low-error regions in the loss landscape, which allows reliable model weight merging.

\subsection{Hierarchical Linear Layer Agreement}
\label{section:layers_agreement}

We propose a hierarchical linear layer agreement mechanism to reliably detect out-of-distribution (OOD) samples and mitigate model misspecification from noisy data. For detected OOD samples, adaptation (backpropagation) is skipped, and the selected layer \(\bphi^*\) is used solely for inference. In contrast, in-distribution (ID) target batches undergo both inference and adaptation (see also Fig.~\ref{fig2:arch}). 
\bluett{Notably, skipping of adaptation on OOD samples aligns with the procedure established in EATA~\cite{niu2022efficient} and SAR~\cite{niu2022towards}, which have been proven that adapting to OOD data harms performance during test-time.}

We measure layer agreement via mutual information between the logits of \(\bphi^*\) and other layers in \(\Phi \setminus \{\bphi^*\}\), quantifying shared information to assess consistency. This allows precise OOD identification by detecting discrepancies in hierarchical representations under shifts. For a target sample \(\mathbf{x}_t\), let \(\mathbf{p}_{\bphi^*}\) and \(\mathbf{p}_{\bphi}\) denote the softmax logits of \(\bphi^*\) and another layer \(\bphi \in \Phi \setminus \{\bphi^*\}\), respectively. Their mutual information~\cite{kraskov2004estimating} is:
\begin{equation}
\label{eq:info}
I(\mathbf{p}_{\bphi^*}; \mathbf{p}_{\bphi}) = H(\mathbf{p}_{\bphi^*}) - H(\mathbf{p}_{\bphi^*} \mid \mathbf{p}_{\bphi})
\end{equation}
where \(H(\cdot)\) is entropy and \(H(\cdot \mid \cdot)\) is conditional entropy.

The average mutual information across layers is then computed as:
\begin{equation}
\label{eq:Iavg}
I_{\text{avg}} = \frac{1}{|\Phi \setminus \{\bphi^*\}|} \sum_{\bphi \in \Phi \setminus \{\bphi^*\}} I(\mathbf{p}_{\bphi^*}; \mathbf{p}_{\bphi})
\end{equation}

If \(I_{\text{avg}} < \tau_{\text{OOD}}\) (where \(\tau_{\text{OOD}}\) is a predefined threshold), the sample is classified as OOD, prompting inference-only mode. Otherwise, it is treated as ID, signaling adaptation. This efficient mechanism allows Hi-Vec to robustly handle OOD noise samples while ensuring targeted updates for ID samples under diverse distribution shifts.

\noindent \textbf{Algorithms.} {We provide the detailed algorithm for test-time adaptation below in Algorithm~\ref{alg:test_time_adaptation}. For source training, we provide the algorithm in the supplement. }

\begin{algorithm}[ht!]
\small
\caption{Test-Time Adaptation with Hi-Vec
{\textbf{Input:}} Unlabeled target domain \(\mathcal{T}\) with \(N_t\) samples \(\{\x_t\}\); Hierarchical layers \(\Phi = \{\bphi_1, \dots, \bphi_k\}\); Source encoder \(\btheta\); Scaling \(\alpha\); Cosine threshold \(\tau\); MI threshold \(\tau_{\text{OOD}}\)\\
\textbf{Output:} Adapted encoder \(\btheta\) and layers \(\phi \in \Phi\)}
\label{alg:test_time_adaptation}
\begin{algorithmic}[1]
\FOR{\textit{iter} in \((N_t/\mathcal{B}_{te})\)}
    \STATE Sample batch \(\{\x_t^{(k)}\}_{k=1}^{\mathcal{B}_{te}} \sim \mathcal{T}\) \\
    \STATE Compute \(\mathbf{z}_t = f_{\btheta}(\x_t)\)
    \STATE Dynamic selection (Eqn.~\ref{eq:gradient})\\
    \STATE Compute Hierarchical layer agreement (Eqn.~\ref{eq:Iavg})
    \IF{\(I_{\text{avg}} < \tau_{\text{OOD}}\)}
        \STATE {Skip adaptation;} Only inference with  \(\bphi^*,\btheta\)
    \ELSE
        \STATE Update \(\btheta, W_{\bphi^*}\) via \(\mathcal{L}\):\\
            {\raggedright\noindent
            $
            \btheta \gets \btheta - \eta\,\nabla_{\btheta}\mathcal{L}(f_{\btheta}, W_{\bphi^*}, \x_t)
            $
            \\
            {\raggedright\noindent
            $
            W_{\bphi^*} \gets W_{\bphi^*} - \eta\,\nabla_{W_{\bphi^*}}\mathcal{L}(f_{\btheta}, W_{\bphi^*}, \x_t)
            $
            }}
        \STATE For all \(\bphi \in \Phi\), set task vectors: \(\mathbf{v}_{\bphi} \gets W_{\bphi}\)
        \STATE Compute cosine similarities between the weights
        \STATE Merge weights for target information sharing (Eqn.~\ref{eq:update-rule})
        \STATE Predict \(\bphi^*\):
        {\raggedright\noindent
        $
        p(\y_t \mid \x_t) = f_{\btheta}(\x_t; W_{\bphi^*}).
        $}
    \ENDIF
    
\ENDFOR
\end{algorithmic}
\end{algorithm}

\section{Experiments and Results}
\label{sec:exps}
We evaluate Hi-Vec across diverse distribution shifts and integrate it with four state-of-the-art baselines, focusing on two challenging scenarios: {(i)} test-time adaptation with outlier datasets~\cite{yu2024stamp} and {(ii)} spurious correlation datasets~\cite{lee2024entropy}.

\noindent \textbf{Datasets and shifts: Five target datasets and six outlier datasets.} The respective source models are trained on public datasets, i.e. Cifar-10~\cite{krizhevsky2009learning}, Cifar-100~\cite{krizhevsky2009learning}, Waterbirds~\cite{sagawa2019distributionally} and ImageNet~\cite{deng2009imagenet}.
Evaluation is performed with diverse shifts as provided in ~\cite{yu2024stamp,lee2024entropy,hendrycks2019benchmarking}. In the outlier-aware scenario, test batches combine target data from respective corrupted benchmarks Cifar-10-C~\cite{hendrycks2019benchmarking}, Cifar-100-C~\cite{hendrycks2019benchmarking}, or ImageNet-C~\cite{hendrycks2019benchmarking} with outlier samples from LSUN-C~\cite{yu2024stamp}, SVHN-C~\cite{yu2024stamp}, TinyImageNet-C~\cite{hendrycks2019benchmarking}, Places-365-C~\cite{yu2024stamp}, and Textures-C~\cite{yu2024stamp}. For spurious correlation scenarios, we use domain-specific datasets such as Waterbirds~\cite{sagawa2019distributionally} and ColoredMNIST, with non-causal correlations. We provide additional dataset details in the Supplement.

\noindent \textbf{Implementation Details.}  
Hierarchical linear layers are integrated with each encoder, mapping features into subspaces of increasing dimensionality starting from 8 up to the encoder's output dimension (i.e., $m_i = 2^{i+2}$)~\cite{kusupati2022matryoshka}; 512 for ResNet-18 or 2048 for ResNet-50). Training these layers is straightforward and requires no additional modifications. We evaluate Hi-Vec using the same metrics from~\cite{yu2024stamp,lee2024entropy} and adopting their reported baseline performances.  For Hi-Vec, we compute gradient norms with PyTorch's built-in functions and apply its padding utilities to equalize model weights for similarity calculations. \purple{For test-time online adaptation, we adhere to the standard protocol by incrementing the target data consisting of outliers or spurious correlations iteratively and keep updating and evaluating the model following prior works~\cite{yu2024stamp,niu2022efficient,wang2021tent,lee2024entropy}.} \bluett{The direct integration of MRL~\cite{kusupati2022matryoshka} linear layers with baseline TTA methods is not feasible due to their multi-head structure.} Consequently, in the tables, we additionally report vanilla MRL~\cite{kusupati2022matryoshka} as a zero-shot baseline akin to source model performance, selecting the dimension with the highest accuracy for brevity. Hyperparameters and further details are provided in the Supplement. We will release the code publicly.

\noindent \textbf{Integration with State-of-the-Art Test-Time Adaptation Methods.}  
We integrate Hi-Vec with four state-of-the-art baselines: Stamp~\cite{yu2024stamp}, which employs stable memory with self-weighted entropy; Deyo~\cite{lee2024entropy}, focusing on object disentanglement with sample re-weighting; Sar~\cite{niu2023towards}, using sharpness-aware entropy minimization; and Tent~\cite{wang2021tent}, based on entropy minimization.

\subsection{Results for Adaptation with Outlier Datasets}
\label{Section:Outlier}
\blue{In this section, we demonstrate the performance of Hi-Vec in handling adaptation with outlier datasets.} We follow the setup and evaluation protocols from \cite{yu2024stamp}. Specifically, we report three key metrics: Accuracy (ACC), which measures classification correctness; Area under the ROC Curve (AUC), which evaluates performance across all thresholds; and H-Score, calculated as the harmonic mean of ACC and AUC. Higher values are better for these metrics.

Table~\ref{table:cifar_tab1} highlights the constraints of baseline methods, including Tent, Sar, and Stamp, on Cifar-10-C and Cifar-100-C with outlier datasets. Hi-Vec's integration addresses this due to multi-scale linear layers that promote enhanced feature expressivity and dynamic layer selection for adaptive fine-tuning, yielding consistent gains across all metrics. On Cifar-10-C datasets with the four outliers, Hi-Vec's integrations improve the mean accuracy for all methods by 2.47\%. For instance, Hi-Vec + Stamp delivers the highest improvements of +5.7\% ACC (to 83.6\%), +8.2\% AUC (to 91.4\%), and +7.2\% H-Score (to 87.3\%) on Cifar-10-C with Noise; +3.4\% ACC (to 85.7\%) with SVHN-C; and +3.9\% ACC (to 86.5\%) with TinyImageNet-C. Similar improvements hold for Cifar-100-C, where Hi-Vec integrations mostly emerge as top performers.
Next, we also provide results on large-scale datasets, showcasing Hi-Vec's effectiveness on ImageNet using Textures and Places-365 as outlier target datasets. Table~\ref{table:imagenet_new} presents results across 15 different image corruptions as in ~\cite{hendrycks2019benchmarking}. Leveraging its adaptive mechanisms, Hi-Vec enables baseline approaches to advance their performance even further, thereby addressing large-scale outlier datasets.

\begin{table*}[t]
\centering
\small

\resizebox{0.9\textwidth}{!}{
\begin{tabular}{llcccccccccccc}
\toprule
                   &  & \multicolumn{3}{c}{\textbf{Noise}} & \multicolumn{3}{c}{\textbf{SVHN-C}} & \multicolumn{3}{c}{\textbf{LSUN-C}} & \multicolumn{3}{c}{\textbf{TinyImageNet-C}} \\ 
                   \cmidrule(r){3-5} \cmidrule(r){6-8} \cmidrule(r){9-11} \cmidrule(r){12-14}
                   &  \textbf{Methods} &  ACC     & AUC      &H-score      &       ACC     & AUC      &H-score      &        ACC     & AUC      &H-score      &     ACC     & AUC      &H-score      \\ \midrule
\multirow{10}{*}{\rotatebox{90}{\textbf{Cifar-10-C}}} & Source  &  57.3     & 70.4      &  62.3    &  57.3     &   67.4    &  61.1    &  57.3     &  62.8     &  59.6   &57.3       & 64.5     &59.4     \\

&  \bluet{MRL~\cite{kusupati2022matryoshka}}  &  59.7    &  65.7    & 62.6     &  59.7    &   64.4   & 62.0  &  59.7 &   61.9   & 60.2   & 59.7      & 64.1    & 61.8    \\

                   & BN Stats~\cite{nado2020evaluating}  &  72.9     & 68.6      & 70.6     &  78.7     &  75.3     & 76.9     & 79.4      & 79.4      &  79.4    &  79.0     & 72.9      & 75.8     \\

                   & EATA~\cite{niu2022efficient} &   72.9    &  68.5     &70.6      &78.8       & 75.3      & 76.9     &  79.4     &  79.4     &  79.4    &   78.9    &   73.1    &  75.9    \\

                   & CoTTA~\cite{wang2022continual} &  77.3     &  62.4     &  67.3    &    81.6   &    78.6   &  80.1    & 82.2      &     84.2  &  83.2    &   81.9    &   \textbf{75.3}    &  78.4    \\

                   & RoTTA~\cite{yuan2023robust}    &   77.6    &  74.3     &   75.6   &    78.4   &   76.0    &  77.2    &   78.8    &    79.5   & 79.1     &    78.6   &  73.3     &    75.8  \\

                   & SoTTA~\cite{gong2023sotta} &  77.8     &51.7       &61.6      &79.3       &72.8       &75.9      &79.8       & 77.9      & 78.8     & 79.6      & 72.6      & 75.9     \\

                   & OWTTT~\cite{li2023robustness} &  62.3     &  64.4     &  58.5    &    66.1   &   75.3    &  69.6    &    63.1   &        78.9    &   68.5    &    56.3   & 58.8     &  56.2     \\
                   \cmidrule(lr){2-14}
                 \cellcolor{white}
                   & Tent~\cite{wang2021tent} &  77.4     &  48.7     &  59.7    &     80.8  &   54.9    &   65.1   &    81.2   &     62.3  &   70.4   &   81.1    &  65.6     &    72.4  \\

                   & SAR~\cite{niu2022towards} &    72.9   &  68.5      &70.6      & 78.7       & 75.3      & 76.9     &  79.4     &  79.4    &79.4 & 79.0     & 72.9      &   75.8    \\

                   & STAMP~~\cite{yu2024stamp}&   {77.9}    &    {83.2}   &   {80.1}   &     {82.3}  &   {79.2}    &  {80.6}    &   {83.5}    &    {86.3}   &  {84.8}    &   {82.6}    &  74.9     &   {78.5}   \\ 

                    \cmidrule(lr){2-14}
                     \rowcolor{lightblue}
                   &  \textit{Tent + {\textbf{Hi-Vec}}} &   \inc{80.7}    &    \inc{63.0}   &   \inc{70.5}   &     \inc{81.7}  &   \inc{55.4}    &  \inc{66.0}    &   \inc{81.7}    &    \inc{62.8}   &  \inc{71.0}    &   \inc{82.5}    &  \inc{65.8}     &   \inc{73.2}   \\ 
                   
                   \rowcolor{lightblue}
                   &  \textit{SAR + {\textbf{Hi-Vec}}} &   \inc{77.7}    &    \inc{63.8}   &   \inc{70.7}   &     \inc{82.5}  &   \inc{73.3}    &  \inc{77.7}    &   \inc{80.2}    &    \inc{79.9}   &  \inc{80.0}    &   \inc{83.0}    &  \inc{69.7}     &   \inc{75.7}   \\

                   \rowcolor{lightblue}
                   &  \textit{STAMP + {\textbf{Hi-Vec}}} &   \inc{\textbf{83.6}}    &    \inc{\textbf{91.4}}   &   \inc{\textbf{87.3}}   &     \inc{\textbf{85.7}}  &   \inc{\textbf{82.7}}    &  \inc{\textbf{84.2}}    &   \inc{\textbf{84.3}}    &    \inc{\textbf{86.9}}   & \inc{\textbf{ 85.5 }}   &   \inc{\textbf{86.5}}    &  \inc{\textbf{81.1}}     &   \inc{\textbf{83.7}}   \\ \midrule

\multirow{10}{*}{\rotatebox{90}{\textbf{Cifar-100-C}}} & Source  &  35.8     &   43.1    &  38.0    &    35.8   &  49.4     &  40.1    &  35.8     &    58.2   &   43.2   &    35.8   &  57.1     & 42.7     \\

&  \bluet{MRL~\cite{kusupati2022matryoshka}} &  41.4    & 60.3     &  47.8   &  41.4     &   61.0   & 47.3  &  41.4 & 58.0     &  48.3   & 41.4      & 63.3    & 48.4    \\

                   & BN Stats~\cite{nado2020evaluating} &  45.8     &    80.9   &   58.4   &  52.7     &   72.5    &   60.9   &  53.7     &   73.8    &  62.0    &    53.2   &   68.6    &   59.7   \\

                   & EATA~\cite{niu2022efficient}  &    55.2   &  86.1     &   67.1   &     58.1  &   75.6    &  65.6    &  58.8    &    77.2   &   66.7   &    58.6   &   70.7    &    64.0  \\

                   & CoTTA~\cite{wang2022continual} &   47.0    & 83.4      &  59.9    &     53.7  &   73.2    &    61.8  &  54.3     &    76.9   &  63.6    &  54.5     &   68.1    &    60.4  \\

                   & RoTTA~\cite{yuan2023robust} &   47.9    &   54.0    &  49.4    &     47.3  &   67.0    &   55.3   &  48.3     &     69.5  &   56.7   &    47.8   &    65.5   &    55.0  \\

                   & SoTTA~\cite{gong2023sotta} &   54.4    &  53.3     & 52.8     &     53.6  &    70.3   &  60.7    &   54.4    &     70.8  &   61.4   &     53.9  &    68.4   &   60.1   \\

                   & OWTTT~\cite{li2023robustness} &  47.1     &  70.3     &   56.2   &      53.9 &    74.3   &   62.3   &   54.5    &      73.5 & 62.5     &   54.2    &    68.5   &   60.4   \\
                   \cmidrule(lr){2-14}
                   \cellcolor{white}  & Tent~\cite{wang2021tent}&   47.9    &   55.8    &   51.2   &  54.4     &  70.4     &  61.2    &    55.4   &    72.4   & 62.7     &   55.0    &    68.6   & 60.9     \\

                   & SAR~\cite{niu2022towards} &   57.5    &    88.6   &  68.9    &     59.2  &  65.2     &   61.9   &   60.5    &    73.5   &   66.3   &     60.8  &   72.1    & 65.9      \\

                   & STAMP~~\cite{yu2024stamp}&   {57.9}    & {98.4}      & {72.8}     &     {63.7}  &   {82.1}    &   {71.7}   &  {63.7}     &     \textbf{82.6}  &  {71.9}    &   {63.9}    &   {75.5}    &    {69.2}  \\ 
                \cmidrule(lr){2-14}
                \rowcolor{lightblue}  &  \textit{Tent + \textbf{Hi-Vec}} &   \inc{54.9}    & \inc{68.2}      & \inc{60.1}     &     \inc{54.7}  &   \inc{73.9}   &   \inc{62.3}   &  \inc{57.1}     &     \inc{72.7}  &  \inc{63.9}    &  \inc{55.3}    &   \inc{69.9}    &   \inc{61.1} \\ 
                
                 \rowcolor{lightblue}  &  \textit{SAR + \textbf{Hi-Vec}} &   \inc{57.9}    & \inc{89.2}      & \inc{69.4}     &     \inc{54.9}  &   \inc{73.4}   &   \inc{62.8}   &  \inc{60.9}     &     \inc{73.9}  &  \inc{66.7}    &  \inc{62.1}    &   \inc{74.4}    &   \inc{68.4} \\ 

                \rowcolor{lightblue} &  \textit{STAMP + \textbf{Hi-Vec}} &   \inc{\textbf{58.2}}    & \inc{\textbf{89.6}}      & \inc{\textbf{73.5}}     &     \inc{\textbf{64.4}}  &   \inc{\textbf{82.5}}   & \inc{\textbf{72.1  }}   &  \inc{\textbf{63.8}}     &     {82.5}  & \inc{\textbf{ 72.0  }}  &  \inc{\textbf{ 64.6  }}  & \inc{\textbf{ 75.8 }}    &  \inc{\textbf{ 70.4}} \\ 

\bottomrule

\end{tabular}
}

\caption{\textbf{Results on adaptation with outlier datasets} using Cifar-10-C and Cifar-100-C as target datasets with four outlier datasets. We report the baselines and use evaluation metrics as provided by~\citet{yu2024stamp} with ResNet-18. Our results are averaged over five runs.  Hi-Vec consistently improves performance (indicated by \incsymb{  } ) of all methods integrated with. Hi-Vec integrations emerge as top performers (\textbf{bold}) }
\label{table:cifar_tab1}
\end{table*}

\begin{table*}[t]
\vspace{-3mm}
\centering
\begin{minipage}[t]{0.57\textwidth}
\centering
\tiny
\resizebox{\linewidth}{!}{%
\begin{tabular}{lcccccc}
\toprule
 & \multicolumn{3}{c}{\textbf{Places365-C}} & \multicolumn{3}{c}{\textbf{Textures-C}} \\ 
\cmidrule(r){2-4} \cmidrule(r){5-7}
\textbf{Methods} & ACC & AUC & H-score & ACC & AUC & H-score \\ \midrule
Source    & 18.2  & 61.6  & 26.1  & 18.2  & 54.6 & 25.8 \\
\bluet{MRL~\cite{kusupati2022matryoshka}}    & 18.4 & 61.9  & 28.3  & 18.4  & 54.6 & 27.4 \\

BN Stats~\cite{nado2020evaluating} & 31.1  & 67.7  & 41.1  & 31.6  & 61.2 & 40.7 \\
EATA~\cite{niu2022efficient} & {46.4}  & 72.6  & 56.0  & 46.4  & 62.2 & 52.8 \\
CoTTA~\cite{wang2022continual} & 33.8  & 66.9  & 43.5  & 34.2  & 60.7 & 42.8 \\
RoTTA~\cite{yuan2023robust} & 36.6  & 68.6  & 46.5  & 37.0  & 65.3 & 46.5 \\
SoTTA~\cite{gong2023sotta} & 41.7  & 67.8  & 50.7  & 41.8  & 60.3 & 48.8 \\
OWTTT~\cite{li2023robustness} & 9.1  & 54.0  & 13.9  & 9.4   & 59.4 & 14.6 \\
\midrule
Tent~\cite{wang2021tent} & 34.9  & 51.8  & 39.5  & 39.0  & 48.6 & 42.0 \\
SAR~\cite{niu2022towards} & 44.9  & 73.3  & 55.0  & 45.6  & 67.0 & 54.0 \\
STAMP~\cite{yu2024stamp} & 46.4  & 77.7  & 57.6  & 46.5  & 71.9 & 56.2 \\
\midrule
\rowcolor{lightblue}
\textit{Tent + \textbf{Hi-Vec}} & \inc{35.4}     & \inc{52.0}    & \inc{42.1}    & \inc{39.4}     & \inc{59.1}    & \inc{47.2} \\
\rowcolor{lightblue}
\textit{SAR + \textbf{Hi-Vec}}  & \inc{45.3}    & \inc{73.8}     & \inc{56.1}    & \inc{46.2}     & \inc{67.4}    & \inc{54.8} \\
\rowcolor{lightblue}
\textit{STAMP + \textbf{Hi-Vec}} & \inc{\textbf{46.9}}    & \inc{\textbf{77.9}}   & \inc{\textbf{58.5}}   & \inc{\textbf{46.8}}    & \inc{\textbf{71.7}}   & \inc{\textbf{56.6}} \\
\bottomrule
\end{tabular}%
}
\caption{\textbf{Results on adaptation with outlier datasets} using Imagenet-C with Places365-C and Textures-C as outlier datasets with ResNet-50. We report baselines and use evaluation metrics as provided by~\cite{yu2024stamp}. Our results are averaged over five runs. Hi-Vec integrations offer improvement (\incsymb{  } ) and are the top performers (bold).
}
\label{table:imagenet_new}
\end{minipage}
\hfill
\begin{minipage}[t]{0.375\textwidth}
\centering
\tiny %
\resizebox{\linewidth}{!}{%
\begin{tabular}{clcc}
\toprule
\textbf{Dataset} & \textbf{Methods}  & \textbf{ACC (\%)} & \textbf{Worst-Group ACC (\%)}  \\
\midrule
\multirow{9}{*}{\rotatebox{90}{\textbf{ColoredMNIST}}} 
    & Source                              & 63.40   & 20.05  \\
     & \bluet{MRL~\cite{kusupati2022matryoshka}}                              & 65.13   & 21.02 \\
    & Tent~\cite{wang2021tent}             & 57.06   & 9.80   \\
    & MEMO~\cite{zhang2021memo}             & 63.77   & 6.23   \\
    & SENTRY~\cite{prabhu2021sentry}         & 63.23   & 15.78  \\
    & EATA~\cite{niu2022efficient}          & 60.81   & 17.98  \\
    \cmidrule(lr){2-4}
    & SAR~\cite{niu2023towards}             & 58.37   & 12.36  \\
    & Deyo~\cite{lee2024entropy}            & 78.24   & 67.39  \\
    \cmidrule(lr){2-4}
    \rowcolor{lightblue}
    & \textit{SAR + \textbf{Hi-Vec}}                & \inc{62.71}   & \inc{15.68}  \\
    \rowcolor{lightblue}
    & \textit{Deyo + \textbf{Hi-Vec}}                & \inc{\textbf{79.53}}   & \inc{\textbf{68.62}}   \\
\midrule
\multirow{9}{*}{\rotatebox{90}{\textbf{WaterBirds}}}
    & Source                              & 83.16   & 64.90  \\
    & \bluet{MRL~\cite{kusupati2022matryoshka}}                              & 85.24  & 60.31 \\
    & Tent~\cite{wang2021tent}             & 82.95   & 54.14  \\
    & MEMO~\cite{zhang2021memo}             & 82.34   & 50.47  \\
    & SENTRY~\cite{prabhu2021sentry}         & 85.77   & 60.90  \\
    & EATA~\cite{niu2022efficient}          & 82.38   & 52.38  \\
    \cmidrule(lr){2-4}
    & SAR~\cite{niu2023towards}             & 82.60   & 53.41  \\
    & Deyo~\cite{lee2024entropy}            & 87.42   & 73.92  \\ 
    \cmidrule(lr){2-4}
    \rowcolor{lightblue}
    & \textit{SAR + {\textbf{Hi-Vec}}}                & \inc{83.25}   & \inc{55.61}  \\ 
    \rowcolor{lightblue}
    & \textit{Deyo + \textbf{Hi-Vec}}                & \inc{\textbf{89.53}}   & \inc{\textbf{77.23}}   \\ 
\bottomrule
\end{tabular}%
}
\caption{\textbf{Results for spurious correlation datasets} using ColoredMNIST and WaterBirds with ResNet-18. We report baselines and metrics by~\cite{lee2024entropy}. Our results are averaged over five runs. Hi-Vec improves the baseline methods and performs the best ({bold}).   }
\label{tab:waterbirds_new}
\end{minipage}
\end{table*}

Additionally, in Supplement~\textcolor{wacvblue}{B1}, we demonstrate Hi-Vec's effectiveness for open-set test-time adaptation, where unseen categories appear at test time but are absent during source training. Moreover, in Supplement~\textcolor{wacvblue}{B2}, we evaluate Hi-Vec under the standard test-time adaptation setting for CIFAR-10-C and CIFAR-100-C~\cite{wang2021tent,niu2022efficient}, that is, without outliers. Across both scenarios, Hi-Vec's integrations with baseline achieve consistent accuracy gains compared to using them without our framework, and are the top performers in their respective experimental conditions.

\subsection{Results for Spurious Correlations Datasets}
\label{Section:Spurious}

\noindent Similarly, Table~\ref{tab:waterbirds_new} highlights the constraints of baseline methods in addressing spurious correlation shifts on the Waterbirds and Colored-MNIST datasets from~\cite{lee2024entropy}, where performance often degrades due to misleading correlations in worst-group subgroups. We evaluate adaptation performance using accuracy, which measures overall classification correctness, and worst-group accuracy, assessing robustness in the most challenging subgroups. Integration with Hi-Vec overcomes the constraints of baseline methods, thereby achieving consistent gains for the integrations across baselines. For instance, Hi-Vec+DeYo delivers the highest improvements on Waterbirds, while Hi-Vec+Sar achieves notable advances on Colored-MNIST, often achieving top performance (in bold). Overall, Hi-Vec's integration improves robustness against spurious correlations, often achieving top performance.

\subsection{Further Benefits and Insights}
\label{Section:benefits}

\begin{figure*}[ht!]
\centering 
\centerline{   
    \includegraphics[width=0.66\columnwidth]{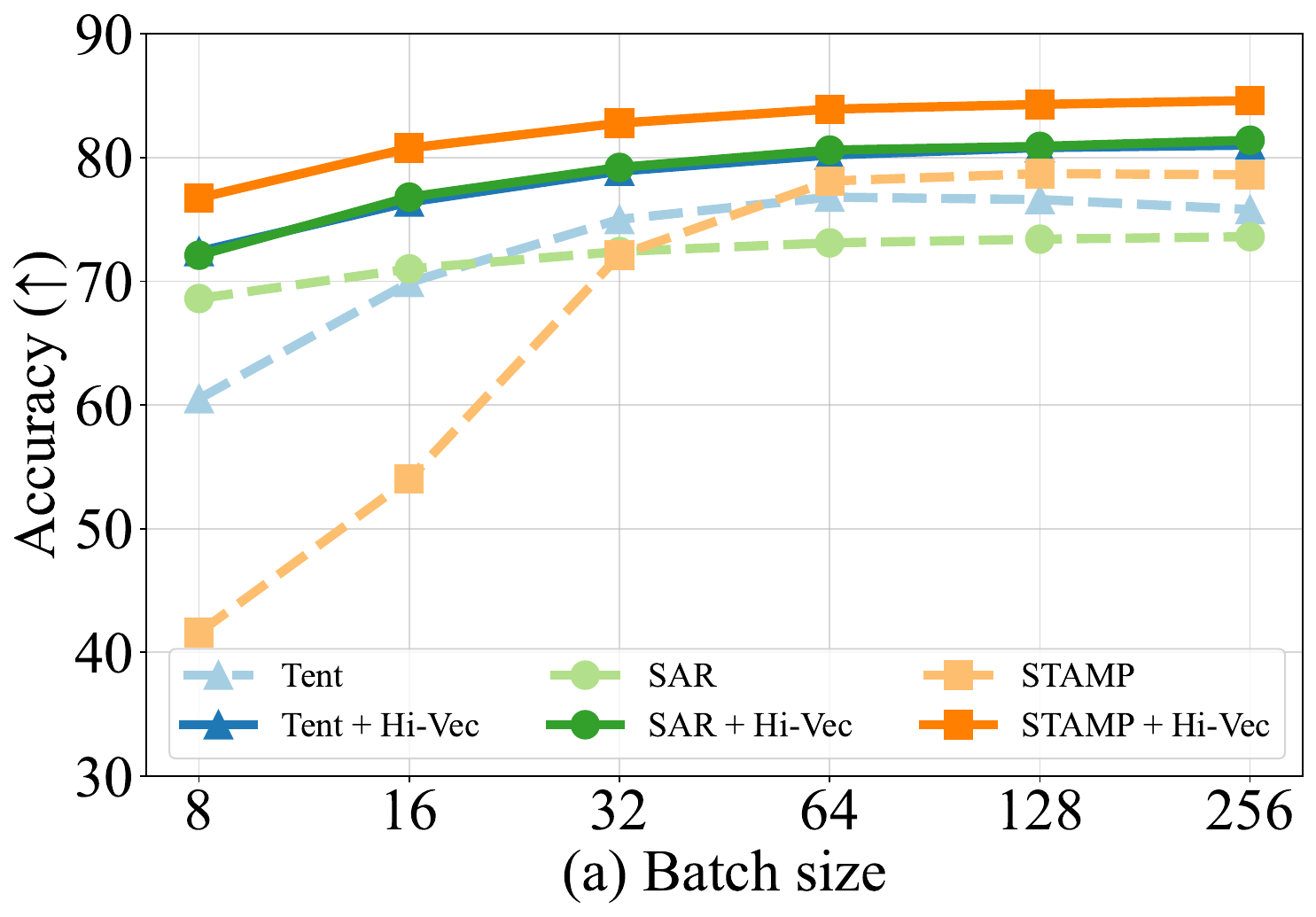} ~
    \includegraphics[width=0.66\columnwidth]{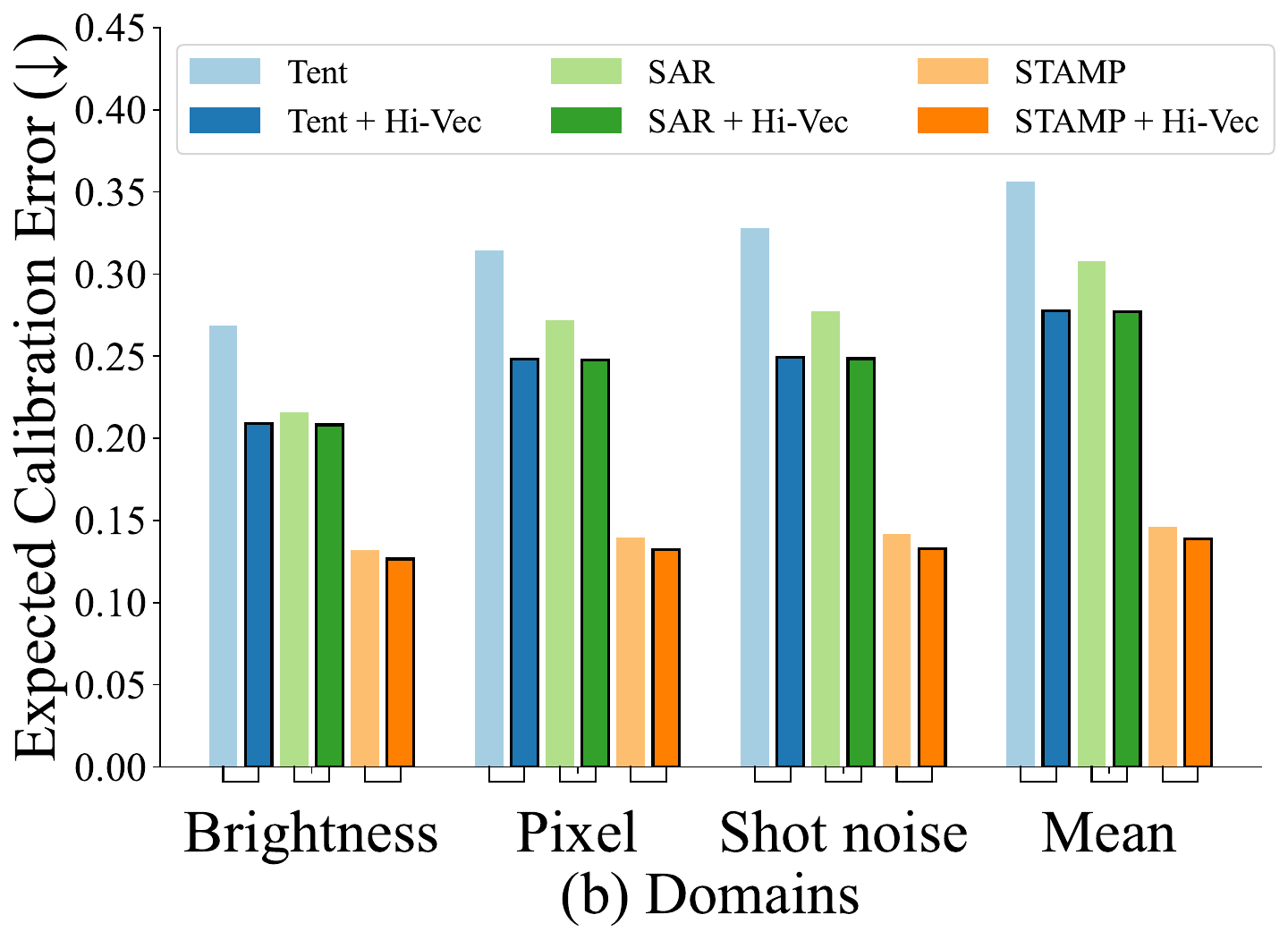} ~
    \includegraphics[width=0.66\columnwidth]{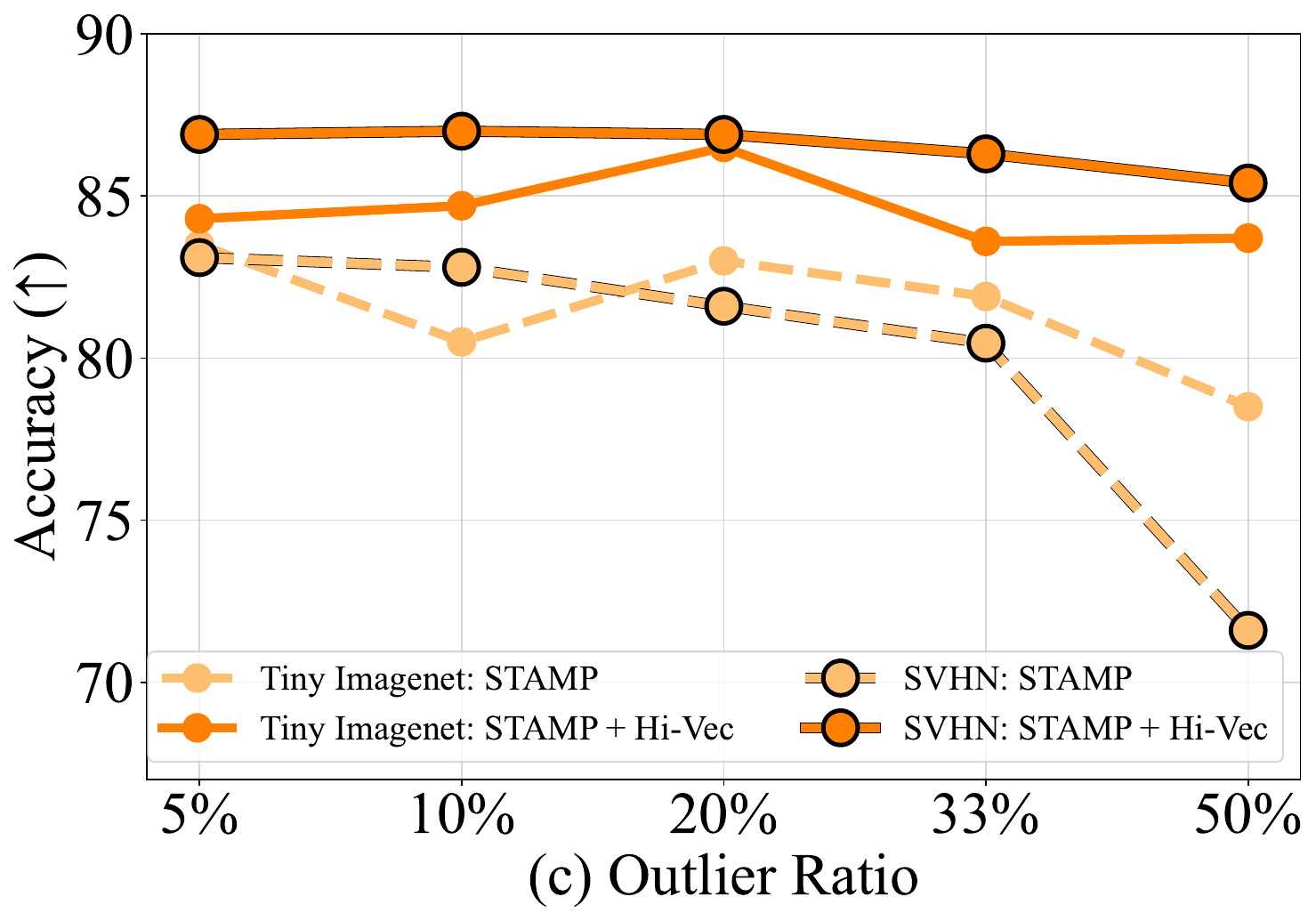} ~
} 
\caption{\textbf{Hi-Vec offers additional benefits: (a) Handles small batch sizes, (b) Addresses uncertainty, (c) Robust to increased outliers.} Hi-Vec performs well over all the common methods with small batch sizes, improving common adaptation methods to handle scenarios reflective of real-world conditions. Moreover, Hi-Vec also improves addressing uncertainty (lower is better) and a higher proportion of outliers in test batches, ensuring robust performance.
}
\label{fig:additional_exps}  
\end{figure*}

\begin{table}[h]
\centering
\resizebox{0.7\columnwidth}{!}{
\begin{tabular}{lcccc}
\toprule
\textbf{ } & \textbf{DS} & \textbf{TV} & \textbf{HLA} & \textbf{Cifar-10-C} \\
\midrule
Source  &  &  &  & 57.3 \\
\bluet{MRL~\cite{kusupati2022matryoshka}}  &  &  &  & 59.7\\
\midrule 
DS         & \checkmark &  &  & 81.2 \\
DS + TV    & \checkmark & \checkmark &  & 82.4 \\
\rowcolor{lightblue}
 \textit{\textbf{Hi-Vec} + Stamp} & \checkmark & \checkmark & \checkmark & \textbf{83.6} \\
\bottomrule
\end{tabular}
}
\vspace{-2mm}
\caption{\textbf{Benefits of all the components} for Cifar-10-C dataset with outliers using ResNet-18. We ablate all the components: Dynamic Selection (DS), Task Vectors (TV), and Hierarchical Layers Agreement (HLA). Each component increases the performance.}
\label{tab:ablation_3components}
\vspace{-4mm}
\end{table}

\noindent \textbf{Handles small batch sizes. }  
The baseline often suffers performance degradation with smaller batch sizes, which are frequently encountered in real-world scenarios.
To evaluate the effectiveness of Hi-Vec across varying batch sizes, we compare the performance of three baseline test-time adaptation methods Tent~\cite{wang2021tent}, Sar~\cite{niu2022efficient}, and Stamp~\cite{yu2024stamp} with and without Hi-Vec. We use Cifar-10-C datasets with 20\% noise outliers and ResNet-18 at test-time. As shown in Fig.~\ref{fig:additional_exps}a, the performance of all three baseline methods decreases significantly as batch sizes are reduced from 32 to 16 and 8. In contrast, integration with Hi-Vec leads to a more stable performance, consistently enhancing all three methods across all batch sizes. This highlights Hi-Vec's ability to improve baseline test-time adaptation methods by maintaining robust performance across varying data conditions, including scenarios with limited batch sizes.

\noindent \textbf{Addresses uncertainty. }
Addressing uncertainty is essential for reliable test-time adaptation~\cite{niu2024test, guo2017calibration}.
Expected calibration error~\cite{guo2017calibration} is used to evaluate the alignment between the model's predicted confidence and its actual accuracy. Lower Expected Calibration Error (ECE) values indicate better-calibrated models, which are essential for robust deployment in real-world scenarios. To evaluate Hi-Vec's effectiveness in addressing uncertainty, we compute the Expected Calibration Error (ECE)~\cite{guo2017calibration} for three baseline methods: Tent, Sar, and Stamp, with and without Hi-Vec. We use Cifar-10-C datasets with 20\% noise outliers and ResNet-18 at test-time. ECE~\cite{guo2017calibration} serves as a metric of misalignment between predicted probabilities and true accuracy, highlighting calibration quality across test conditions. As shown in Fig.~\ref{fig:additional_exps}b, all baseline methods exhibit higher ECE values, indicating that they are not well calibrated for distribution shifts. In contrast, integrating Hi-Vec with these baseline methods significantly reduces ECE error across all the domains. This improvement can be attributed to Hi-Vec's hierarchical structure and classifier agreement mechanism, which selectively skips adaptation for noisy outlier batches, thereby preventing error accumulation.

\noindent \textbf{Robust to increased outliers.}  
The baseline methods often struggle with high proportions of outliers in the target data, leading to significant performance degradation. Hi-Vec's integration for baseline methods improves adaptation robustness to outliers. To demonstrate this, following the setting of ~\cite{yu2024stamp}, we conduct experiments with two outlier datasets, SVHN and TinyImageNet, with ResNet-18, increasing the proportions of outlier samples introduced at test time from 5\% to 50\%. Fig.~\ref{fig:additional_exps}c illustrates the results with and without Hi-Vec. The accuracy of Stamp decreases significantly with an increasing proportion of outliers, particularly for the SVHN dataset. In contrast, Hi-Vec consistently outperforms Stamp across all outlier ratios, achieving the highest improvements when outlier proportions are large. Furthermore, Hi-Vec's integrations demonstrate minimal degradation in accuracy as the proportion of outliers increases, highlighting its robustness. In real-world scenarios, target datasets often contain varying levels of outliers that can affect the method's performance. Hi-Vec mitigates increased outliers due to its coarse-to-fine representation learning and layer agreement mechanism. This helps to prevent backpropagation on noisy outlier batches, improving robust adaptation.

\noindent \textbf{Benefits of each component. } 
We demonstrate the effectiveness of each component of Hi-Vec by ablating them.
In Table~\ref{tab:ablation_3components}, we demonstrate this for the Cifar-10-C datasets with 20\% noise outliers and ResNet-18 at test-time and Hi-Vec integrated with Stamp. In this table, the baseline source and MRL~\cite{kusupati2022matryoshka} have been evaluated in a zero-shot manner akin to their original protocols.
The introduction of hierarchical linear layers with dynamic layer selection improves the performance significantly than the source and MRL~\cite{kusupati2022matryoshka} baseline. Additionally, target information sharing between hierarchical layers with task vectors enhances the performance further. Finally, with the incorporation of the hierarchical layers agreement, the method achieves the best result, demonstrating the effectiveness of Hi-Vec.   

\noindent \textbf{Additional Benefits and Analysis of Hi-Vec.} We provide detailed inference time comparisons in Supplement~\textcolor{wacvblue}{B3}, showing Hi-Vec achieves robust adaptation with minimal computational overhead. In Supplement~\textcolor{wacvblue}{B4}, we highlight Hi-Vec's ability to mitigate catastrophic forgetting by retaining source information. Supplement~\textcolor{wacvblue}{B5} offers insights into hierarchical layer agreement, while Supplement~\textcolor{wacvblue}{B6} includes Grad-CAM visualizations and layer selection histograms across two target datasets.

\section{Conclusion}  
We proposed \textbf{Hi-Vec}, a novel framework that integrates with state-of-the-art test-time adaptation methods to allow them to address diverse shifts at test-time.
By leveraging hierarchical linear layers, Hi-Vec learns coarse-to-fine representations that can be seamlessly integrated with existing methods.
We further advance the architecture by integrating dynamic layer selection for precise adaptation, target information sharing for cross-layer coherence, and layer agreement to detect noisy samples and skip adaptation on noisy test batches.
We have evaluated our framework, Hi-Vec, in challenging scenarios, including outlier shifts and spurious correlations. Our results demonstrate the strong capabilities of Hi-Vec, consistently enhancing existing methods across diverse shifts. Additionally, our framework significantly improves baseline methods in handling small batch sizes, addressing high outlier samples in target data, mitigating uncertainty, and supporting both standard and open-set test-time adaptation settings, demonstrating its versatility for real-world challenges.   
The strong performance of Hi-Vec motivates us to further enhance its capabilities. \bluett{Notably, eventhough Hi-Vec relies on backpropagation, it can be extended to parameter-free adaptation methods but further research is needed.}

\section*{Acknowledgments}
This work is supported by DAAD programme Konrad Zuse Schools of Excellence in Artificial Intelligence and the Munich Center for Machine Learning, both sponsored by the Federal Ministry of Research, Technology and Space. DML and JAS received funding from HELMHOLTZ IMAGING, a platform of the Helmholtz Information and Data Science Incubator. MH is in part supported by the Munich School of Data Science (MuDS) and the European Laboratory for Learning and Intelligent Systems (ELLIS) PhD program.
LD is supported but the German Federal Ministry of Research, Technology and Space (DECIPHER-M, 01KD2420G).

\appendix

\clearpage
\newpage

\section{Additional discussions}

\noindent \subsection{Rationale for using Hierarchical layers in FC Layers but not in the Encoder}Common convolutional encoders such as ResNets~\citep{he2016deep} learn multiple representations, with initial layers capturing low-level features, and deeper layers learning high-level semantics~\citep{zeiler2014visualizing}. However, standard fully connected (FC) layers attached to the encoder flatten these structured hierarchies into a single vector through the transformation \(\phi(\mathbf{x}) \mapsto \mathbf{W}\phi(\mathbf{x}) + \mathbf{b}\), leading to `information diffusion'~\cite{kusupati2022matryoshka}. This conflicts with the encoder's multi-scale inductive bias, as shown by the rank inequality \(\text{Rank}(\phi_{\text{conv}}) \gg \text{Rank}(\phi_{\text{FC}})\), where \(\phi_{\text{conv}}\) represents convolutional features and \(\phi_{\text{FC}}\) the FC-processed output~\citep{feng2022rank_2}. To address this, matryoshka representation learning~\cite{kusupati2022matryoshka} introduced nested linear projections \(\{f_i\}_{i=1}^k\), where each \(f_i: \mathbb{R}^d \rightarrow \mathbb{R}^{d_i}\) satisfies dimensional constraints \(d_1 < \cdots < d_k = d\), enforced by truncation conditions \(\forall i<j,\ f_i(\phi(\mathbf{x})) = \text{Trunc}_{d_i}(f_j(\phi(\mathbf{x})))\)~\citep{kusupati2022matryoshka}. By using hierarchical linear layers~\cite{kusupati2022matryoshka} exclusively as the classification layers after the encoder, we enable it to learn coarse-to-fine features while preserving the encoder's hierarchical structure. This avoids redundancy and adaptation of already robust encoder features and addresses the limitations of fixed-dimensional FC layers, which are likely to fail to adapt to diverse distribution shifts for test-time adaptation.

\noindent \bluet{\subsection{Modification of Source Training} The test-time adaptation landscape comprises two primary approaches: (i) Fully test-time adaptation methods and (ii) Methods that modify source training (often termed training preparation~\cite{xiao2024beyond}). Fully test-time adaptation methods~\cite{wang2021tent,lee2024entropy} leave source training unaltered, and in line with this, our three proposed innovations are applied exclusively at test time. Common methods, however, often rely on single-dimensional architectures that fail to capture multi-scale learning across the model. Consequently, we enhance common TTA methods to tackle diverse distribution shifts by incorporating hierarchical linear layers during source training. Similarly, other TTA approaches~\cite{xiao2022learning,xiao2023energy,liu2021ttt++,dubey2021adaptive,osowiechi2023tttflow} also modify source training to develop strategic priors for adaptation.}

\noindent \subsection{Training Hierarchical linear layers} Training the hierarchical linear layers during source training is straightforward. All layers are optimized using a standard cross-entropy loss with equal weight scaling across the hierarchical linear layers~\cite{kusupati2022matryoshka}, i.e., requiring no additional optimizations. At test time, our gradient selection mechanism adapts only the layer with the lowest gradient norm per batch, while the others are updated via a sharing mechanism.

\noindent \subsection{Rationale for the usage of gradient norm to select the optimal linear layer for inference and adaptation }  
Gradient norm quantifies the magnitude of alteration that needs to be applied to align the model output with the target distribution. In our case, each hierarchical linear layer learns coarse-to-fine representations. Therefore,  the smallest gradient norm corresponds to the layer whose parameters are already well-aligned to model the target data and require only minor adjustments. Hence, by choosing the layer with the lowest gradient norm for inference and adaptation, our method relies on the most stable features while minimizing the risk of over-adaptation and instability~\cite{wang2021tent,iwasawa2021test}. This criterion promotes efficient test-time adaptation by using the layer's inherent common target features with minimal updates.\\

\noindent \subsection{Motivation for using task vectors to share target information between the linear layers}  
The hierarchical linear layers are designed to learn coarse-to-fine representations of increasing granularity. If only the most suitable linear layer is adapted during test time, while the remaining layers stay unchanged, the non-adapted layers will not incorporate any new information or statistics about the target data. This undermines the utility of learning coarse-to-fine representations, as the non-adapted layers do not reflect the updated target distribution's statistics. To address this, task vectors are employed to propagate target information from the adapted layer to the remaining layers. By leveraging lightweight transformations, task vectors ensure that all layers align with the target distribution without requiring additional backpropagation. This maintains the cohesiveness of the hierarchical structure and allows all layers to contribute meaningfully to robust adaptation under diverse test-time conditions.

\noindent\subsection{Rationale behind using cosine similarity to find similar layers for target information sharing} The weights of hierarchical linear layers have varying dimensionalities, as each layer is designed to learn coarse-to-fine representations. Cosine similarity is used to identify similar layers, eligible for  target information sharing. The metric measures the alignment between weight vectors, focusing on their directional similarity rather than their magnitude. This property makes cosine similarity particularly effective for comparing weights across layers with different scales or norms. Computationally, to handle the dimensional mismatch between weights, smaller weight matrices are padded with zeros to match the largest dimensionality among the layers. This zero-padding aligns all weight vectors in a common vector space without altering their intrinsic directional properties, as the added zeros do not affect the cosine similarity computation~\cite{dieleman2016exploiting}. By leveraging cosine similarity in this manner, we can robustly and efficiently identify layers with similar weights, enabling effective  target information sharing across hierarchical linear layers during test-time adaptation.

\noindent \subsection{Motivation for using mutual information in Hierarchical linear layers Agreement} Mutual information quantifies the dependency between two sets of variables, capturing both linear and nonlinear relationships~\cite{zhao2021domain,guo2025smoothing}. In Hi-Vec, we use mutual information to evaluate the agreement between the logits of the optimal hierarchical layer, $\mathbf{p}_{\bphi^*}$, and those of the remaining layers, $\mathbf{p}_{\bphi}$ for $\bphi \in \Phi \setminus \{\bphi^*\}$. The hierarchical linear layers in Hi-Vec capture coarse-to-fine representations, where each layer learns features at different levels of granularity. These layers are connected sequentially in a nested architecture, transforming the encoder's output vector 'z' to match the dimensionality required by the linear layer. Agreement among their corresponding logits reflects the presence of shared features across these levels. ID samples exhibit high agreement across hierarchical layers as their features align with the source domain's representations. In contrast, OOD samples disrupt this agreement, leading to lower shared information between the logits. By quantifying this inter-layer agreement, mutual information enables Hi-Vec to detect OOD samples and avoid unnecessary adaptation to noisy batches. Moreover, mutual information has proven effective in domain adaptation by aligning features and reducing discrepancies~\cite{zhao2021domain,de2023rule}. For instance, Zhao et al.~\cite{zhao2021domain} use mutual information to align features in unsupervised domain adaptation, while De Bernardi et al.~\cite{de2023rule} leverage it for OOD detection. Hi-Vec uses Mutual information to ensure selective adaptation on ID samples, preserving model stability by avoiding model misspecification and improving computational efficiency during test-time adaptation.

\noindent \subsection{Motivation for not using efficient and fixed-feature versions from Matryoshka representation learning}  
The Matryoshka model~\cite{kusupati2022matryoshka} also includes an efficient version that nests the linear layers within one shared single linear layer. Additionally, the fixed feature version is provided that maps encoder output to one single dimension between the 8 to the highest dimension of the encoder and treats it as one single fc layer. 
However, in our work, we focus on diverse shifts at test time, which requires flexible models or layers capable of capturing varied features across multiple dimensions. The efficient version of~\cite{kusupati2022matryoshka}, due to the nesting of representations within a single fully connected layer, is not useful for adaptation since the adaptation of one layer affects all of the other layers in the nest. Similarly, the fixed-feature version depends on static representations tied to a single dimension, which will also diffuse information~\cite{kusupati2022matryoshka} as in the common setup. Training multiple encoders with such fixed-feature models  to handle diverse shifts is computationally inefficient and lacks the flexibility required for test-time adaptation.

\bluett{\subsection{{Additional Notations for Merging Layer Weights of Small and Large Linear Layers}}}
\label{section:merging}
\bluett{Let \(W_{\phi}\) denote the weight matrix of layer \(\phi\) with dimensions \((m \times n)\), and let \(W_{\phi^*}\) denote the weight matrix of the dynamically selected layer \(\phi^*\) with dimensions \((k \times l)\). These matrices can be written explicitly as,}
\bluett{\begin{equation*}
    W_{\phi} = \left( w_{\phi,ij} \right)_{i=1,\dots,m;\; j=1,\dots,n},
\end{equation*}
\begin{equation*}
    W_{\phi^*} = \left( w_{\phi^*,ij} \right)_{i=1,\dots,k;\; j=1,\dots,l}.
\end{equation*}}
\bluett{For dimension-aware merging, we define the shared dimensions
\begin{equation*}
    g = \max(m, k), \quad h = \min(n, l),
\end{equation*}
where \(\max\) selects the larger dimension and min selects the smallest dimension in respective directions. Then, the weights of \(W_{\phi}\) are projected into the common shape \((g \times h)\), aligned with the indexing of \(W_{\phi^*}\):
\begin{equation*}
    W_{\phi}[\phi^*] := \left( w_{\phi,ij} \right)_{i=1,\dots,g;\; j=1,\dots,h}.
\end{equation*}}

\bluett{\subsection{ Extending to Parameter-free baselines}}
\noindent \bluett{We also extend Hi-Vec to parameter-free baselines such as Laplacian Adjusted Maximum-likelihood Estimation (LAME)~\cite{boudiaf2022parameter}, which adjusts the model's output for target batches. Specifically, we integrate Hi-Vec's dynamic layer selection to perform targeted inference for each batch instead of representations from a static linear layer. Next, leveraging hierarchical layer agreement to decide whether the model's output adjustment with LAME~\cite{boudiaf2022parameter} is necessary.
More specifically, we implement LAME with a RestNet-18 encoder with and without Hi-Vec for the CIFAR-10-C dataset with 20\% outliers, which depicts the standard setting of our ablation studies. We find that with the implementation of Hi-Vec, the standard implementation of LAME can be improved from 68.4\% to 79.7\% accuracy, which is significantly higher than the source baseline of 57.3\%. 
This highlights the ability of Hi-Vec to also integrate with parameter-free methods apart from backpropagation-based methods and motivates us to further investigate this in future work.}\\

\noindent\subsection{Algorithms} 
\blue{We provide the details about source model training in Algorithm~\ref{alg:source_training}. The detailed test-time algorithm is in the main paper. }

\begin{algorithm}[ht!]
\small
\caption{Source Model Training \\
{\textbf{Input:}} $\mathcal{D}_s$: source domain with labeled samples $\{(\x_s, \y_s)\}$; $\btheta$: encoder parameters; \(\Phi = \{\bphi_1, \bphi_2, \dots, \bphi_k\}\): hierarchical linear models; $\{W_m\}_{m \in \mathcal{M}}$: hierarchical linear layer weights where $\mathcal{M} = \{m_1, m_2, \dots, m_k\}$ spans the dimensions.\\
{\textbf{Output:} learned $\theta$ with $\Phi$ hierarchical linear layers. }
}

\label{alg:source_training}
\begin{algorithmic}[1]
\FOR{\textit{epoch} in $1, \dots, N_{\text{epochs}}$}
    \FOR{\textit{batch} $\{\x_s^{(i)}, \y_s^{(i)}\}_{i=1}^{\mathcal{B}_{tr}}$ in $\mathcal{D}_s$}
        \STATE Extract representations: $\mathbf{z} = f_{\btheta}(\x_s)$.
        \STATE Compute predictions for all layers $\bphi \in \Phi_{}$ and corresponding dimensions \(m \in \mathcal{M}\): \\ $ W_{\bphi}^\top \mathbf{z}_m$, where $\mathbf{z}_m$ are the first $m$ dimensions of $\mathbf{z}_s$.
        \STATE Compute cross-entropy loss: \\ $\mathcal{L} = \sum_{\phi \in \Phi} \sum_{ m \in \mathcal{M}} \mathcal{L}_{\text{CE}}( W_{\bphi}^\top \mathbf{z}_{1:m}, \y_s)$.
        \STATE Update $\btheta$ and $\{W_{\bphi}\}_{\bphi \in \Phi}$ as in Eq.2 from main paper.
    \ENDFOR
\ENDFOR
\end{algorithmic}
\end{algorithm}

\section{Additional Experiments}
\label{Appendix:additional_exp}

\subsection{Results across the open-set adaptation}
\label{app:open_set}
Hi-Vec leverages hierarchical representations and task vectors propagation to adapt to unseen domains, making it effective for open-set test-time adaptation. In this open-set setting, a subset of classes is unseen during source training but present in the test set. Following~\cite{yu2024stamp}, we evaluate Hi-Vec on Cifar-10-c and Cifar-100-c. For Cifar-10-c, the source model is trained on 8 classes and evaluated on 2 unseen classes in the corrupted version. Similarly, for Cifar-100-c, the source model is trained on 80 classes and evaluated on 20 unseen classes. Results are given in Table~\ref{table:open_set_cifar}. Open-set~\cite{panareda2017open} adaptation lacks prior class and corresponding sample information during training, posing challenges for common methods. Hi-Vec improves performance across both scenarios by learning coarse-to-fine representations, avoiding overfitting to learned fixed features and corresponding limited classes during training, thus enabling robust adaptation to unseen classes.

\setlength{\tabcolsep}{7.0pt}

\begin{table}[!t]
\centering
\resizebox{0.92\linewidth}{!}{%
\begin{tabular}{lcccccc}
\toprule
 & \multicolumn{3}{c}{\textbf{CIFAR10-C (8:2)}} & \multicolumn{3}{c}{\textbf{CIFAR100-C (80:20)}} \\ 
\cmidrule(r){2-4} \cmidrule(r){5-7}
\textbf{Methods}  & Acc & Auc & H-score & Acc & Auc & H-score \\ \midrule
Source    & 60.6  & 61.5  & 60.0  & 37.1  & 60.6  & 44.2 \\
Source    & 60.6  & 61.5  & 60.0  & 37.1  & 60.6  & 44.2 \\
MRL & 71.3  & 63.8  & 66.4  & 43.4  & 62.2  & 50.3 \\
BN Stats~\cite{nado2020evaluating}  & 79.4  & 66.9  & 72.6  & 55.5  & 66.2  & 60.2 \\
EATA~\cite{niu2022efficient}  & 81.5  & 67.9  & 74.1  & 61.3  & 67.4  & 64.2 \\
CoTTA~\cite{wang2022continual}  & 81.8  & 65.4  & 72.5  & 57.1  & 66.3  & 61.3 \\
RoTTA~\cite{yuan2023robust}  & 79.4  & 62.4  & 69.8  & 50.6  & 63.7  & 56.2 \\
SoTTA~\cite{gong2023sotta}  & 82.7  & 62.7  & 71.3  & 61.4  & 68.2  & 64.6 \\
OWTTT~\cite{li2023robustness}  & 65.9  & 63.5  & 64.4  & 56.2  & 66.0  & 60.6 \\
\midrule
Tent~\cite{wang2021tent}  & 80.1  & 65.9  & 72.3  & 60.4  & 65.6  & 62.8 \\
SAR~\cite{niu2022towards}  & 82.3  & 66.0  & 73.2  & 63.4  & 69.1  & 66.1 \\
STAMP   & 85.0  & 69.4  & 76.4  & 66.0  & 71.2  & 68.5 \\
\midrule
\rowcolor{lightblue}
\textit{Tent + \textbf{Hi-Vec}}  & \inc{83.9}  & \inc{75.0}  & \inc{79.2}  & \inc{60.6}  & \inc{68.2}  & \inc{64.1} \\
\rowcolor{lightblue}
\textit{SAR + \textbf{Hi-Vec}} & \inc{84.1}  & \inc{74.9}  & \inc{79.2}  & \inc{63.8}  & \inc{69.7}  & \inc{66.6} \\
\rowcolor{lightblue}
\textit{STAMP + {\textbf{Hi-Vec}}} & \inc{\textbf{86.5}}  & \inc{\textbf{69.9}}  & \inc{\textbf{77.3}}  & \inc{\textbf{66.8}}  & \inc{\textbf{72.1}}  & \inc{\textbf{69.3}} \\
\bottomrule
\end{tabular}
}
\\[2mm]
\caption{\textbf{Comparisons on adaptation with open-set datasets} using CIFAR10-C-20 and CIFAR100-C-80 as target datasets. We report the baselines provided by~\citet{yu2024stamp} with ResNet-18. Our results are averaged over fine runs. As in the main results, Hi-Vec improves the performance (\incsymb{  } )  and is the top-performer (bold).  }
\label{table:open_set_cifar}
\end{table}

\subsection{Results across common test-time adaptation setting}
\label{app:usual_setting}
Following common test-time adaptation methods~\cite{wang2021tent,niu2023towards,goyal2022test,yu2024stamp}, we also provide results without any outlier datasets at test time. In Table~\ref{table:usual_setting}, we compare the performance of all baselines across the domains of the Cifar-10-c and Cifar-100-c datasets. Hi-Vec achieves the highest accuracy compared to all recent test-time adaptation methods on both datasets.

\setlength{\tabcolsep}{5.0pt}
\begin{table*}[!t]
\centering
\small
\resizebox{0.95\textwidth}{!}{
\begin{tabular}{llm{0.5cm}m{0.5cm}m{0.5cm}m{0.5cm}m{0.5cm}m{0.5cm}m{0.5cm}m{0.5cm}m{0.5cm}m{0.5cm}m{0.5cm}m{0.5cm}m{0.5cm}m{0.5cm}m{0.5cm}c}
\toprule
\multicolumn{2}{c}{Method} & Gauss & Shot & Imp & Defoc & Glass & Rot & Zoom & Snow & Frost & Fog & Bright & Contr & Elast & Pixel & JPEG & \textbf{Avg. } \\ \midrule
\multirow{10}{*}{\rotatebox{90}{CIFAR10-C}} & Source & 28.7	&35.2	&24.2	& 57.3&49.0	&66.4	&64.8	&76.7	&62.9	&73.4	&	90.3&	31.4&	78.8&	46.5&74.7 & 57.3
 \\ 
& BN Stats\cite{nado2020evaluating}  & 70.2	& 72.0&63.6	&87.6	&66.5	&85.9 &86.9	&82.1	&80.5	&84.1	&90.8	&85.4	&77.3 &78.6	&74.3	&79.0
 \\

&CoTTA\cite{wang2022continual} & 76.1&77.6	&73.9	&87.9	&71.9&	86.9&87.5	&82.8	&82.0	&85.2	&90.8	&85.7	&79.3	&80.4	&78.3	&81.8
  \\
& EATA\cite{niu2022efficient} & 76.5	&77.9	&71.7	&89.1	&70.4 &87.4	&{89.0}	&85.2	&84.2	&86.0	&91.5	&{88.4} &79.9	&83.9	&78.5	&82.6
  \\

 &RoTTA\cite{yuan2023robust} & 69.6	&71.0	&62.9	&87.5&	67.0&	85.9&86.9	&82.4	&79.6	&84.7	&91.2	&73.5	&78.3&77.9	&74.9	&78.2
 \\
& SoTTA\cite{gong2023sotta} &75.8	&79.2	&71.5	&{89.4}	&70.6	&{87.8}	&{89.0}	&85.6	&84.0	&87.2	&{92.4}	&87.4	&79.9	&84.5	&79.0	&82.9
  \\

& OWTTT\cite{li2023robustness}  &70.1	&72.0	&63.0	&86.9	&66.5	&85.7	&86.7	&82.7&	80.9&	84.5&91.2	&83.0	&77.9	&76.8	&74.9	&78.9
 \\ 
 \cmidrule(lr){2-18}
 &Tent\cite{wang2021tent} &	76.1& 78.4	&70.5	&87.8	&70.1	&86.8	&87.5	&84.8	&82.0	&85.1	&91.1	&86.6	&79.4&	82.7&78.5 &81.8
 \\
 & SAR\cite{niu2022towards} &70.7	&72.1	&66.8	&87.6	&68.2	&85.9	&86.9	&82.1	&80.5	&84.1	&90.8	&85.4	&77.3&	78.6&74.3 &79.4
 \\
& STAMP &{80.9}	&{82.9}	&{77.2}	&87.6	&{74.9}&	86.6&87.7 &{85.9}&{85.9}&{88.1}&90.4&87.2&{80.3}&{86.7}&{82.6}&{84.3}\\
\cmidrule(lr){2-18}
\rowcolor{lightblue}

&\textit{Tent + \textbf{Hi-Vec} }& 77.7& 79.6	&70.6	&88.3	&72.3	&87.6	&87.9	&85.7	&82.8	&85.7	&91.9	&86.5	&80.2&	82.9&78.9 &\inc{82.5} \\

\rowcolor{lightblue}

& \textit{SAR + \textbf{Hi-Vec}} &78.5 & 80.2 & 69.1 & 91.4 & 73.8 & 88.7 & 90.7 & 88.0 & 88.3 & 89.0 & 93.2 & 91.6 & 80.5 & 86.4 & 79.2 &\inc{84.6}\\

\rowcolor{lightblue}

& \textit{STAMP + \textbf{Hi-Vec}} &\textbf{83.6}	&\textbf{84.7}	&\textbf{77.3}	&\textbf{90.7}	&\textbf{77.3}&	\textbf{88.8}&\textbf{90.7} &\textbf{88.5}&\textbf{89.6}&\textbf{89.1}&\textbf{92.9}&\textbf{90.3}&\textbf{83.4}&\textbf{88.3}&\textbf{83.6}&\inc{\textbf{86.6}}\\

 \\ \midrule
\multirow{10}{*}{\rotatebox{90}{CIFAR100-C}} & Source &12.4	&14.5	&7.2	&36.0 &44.7	&45.1	&45.2	&49.5	&41.6	&36.9	&63.3	&13.2	&57.5	&23.8	&46.6	&35.8
 \\ 
& BN Stats\cite{nado2020evaluating}  &41.1	&41.3 &38.9	&63.1	&51.5	&60.8 &63.8	&51.2	&53.5	&53.2	&64.4	&59.0	&58.4 &58.0&47.6	&53.7
 \\

&CoTTA\cite{wang2022continual} & 47.0&47.8	&45.1	&59.6	&54.0&59.5	&61.3	&53.3	&55.0	&52.9	&62.6	&50.3	&58.1	&61.8	&53.1	&54.8
  \\
& EATA\cite{niu2022efficient} &50.7	&53.5	&48.1	&67.0	&55.6 &64.8	&67.0	&59.1	&59.2	&60.4	&67.7	&63.9 &61.8	&63.3	&54.6	&59.8
  \\

& RoTTA\cite{yuan2023robust} &35.9	&36.6	&33.8	&60.6	&47.1	&57.7	&60.8	&48.0	&42.2	&50.8	&59.2	&32.1	&53.8	&52.3	&44.4	&47.7
  \\
&SoTTA\cite{gong2023sotta} &51.6	&53.8	&47.4	&66.9&	56.9&	65.3&	68.1&	58.9&	60.1&	60.1&	69.4&	63.1&62.3&62.8	&54.8	&60.1
 \\
& OWTTT\cite{li2023robustness}  &41.6	&42.8	&38.8	&63.4	&52.6	&61.6	&64.8	&53.3&	54.8&	54.5&	65.8&	58.4&	60.1&57.6	&49.3	&54.6
 \\
\cmidrule(lr){2-18}
 &Tent\cite{wang2021tent} &52.3	&52.1	&47.7	&66.9	&56.1	&64.3	&65.3	&58.3	&58.7	&60.0	&67.8	&62.1	&61.8&	63.0   &54.5 &59.4
 \\
 & SAR\cite{niu2022towards} &{55.1}	&55.0	&51.2	&68.4	&58.2	&66.0	&67.4	&60.3	&60.8	&61.9	&69.8	&65.5	&63.6&66.2	&56.8 &61.7
 \\
& STAMP\cite{yu2024stamp} &{57.2}	&{58.5}	&{52.8}	&{69.9}	&{61.4}&	{68.1}& {70.1}&{63.3}&{63.9}&{64.8}&{72.2}&{69.9}&{66.5}&{69.2}&{59.0}&{64.4}\\
\cmidrule(lr){2-18}
\rowcolor{lightblue}

\rowcolor{lightblue}
&\textit{Tent + \textbf{Hi-Vec}}&{52.9}	& {53.4}	&{47.9}	&{67.8}	&56.7	&64.9	&65.7	&58.4	&58.9	&61.9	&68.3	&62.7	&62.4&	63.5   &54.9 &\inc{60.2}\\

\rowcolor{lightblue}
& \textit{SAR + \textbf{Hi-Vec}}&{55.7}	&55.3	&51.8	&69.2	&58.5	&66.7 &67.9	&61.5	&61.7	&62.5	&70.5	&66.2	&63.9&66.7	&57.4 &\inc{62.3}\\

\rowcolor{lightblue}
& \textit{STAMP + \textbf{Hi-Vec}} &\textbf{{57.9}}	&\textbf{{60.2}}	&\textbf{{53.5}}	&\textbf{{70.2}}	&\textbf{{61.9}}&	\textbf{{68.9}}& \textbf{{71.4}}&\textbf{{63.9}}&\textbf{{64.3}}&\textbf{{65.4}}&\textbf{{72.6}}&\textbf{{70.4}}&{\textbf{66.9}}&{\textbf{69.7}}&{\textbf{59.3}}&\inc{\textbf{65.1}}\\

 \\ 
 \bottomrule
\end{tabular}
}
\caption{\textbf{Comparisons on common test-time adaptation setting.} for {CIFAR10-C} and {CIFAR100-C}. Integrating Hi-Vec improves the common methods and achives the  the best results (bold) } 
\label{table:usual_setting}
\end{table*}

\subsection{Inference time}
\label{app:inf_time}
During source training, Hi-Vec introduces a small parameter overhead of approximately 5\% compared to standard implementations. \bluett{For example, for ResNet-18, the encoder has approximately 11M parameters while the added layers contribute approximately 0.5M parameters. This overhead scales modestly to larger models, even when implemented for intermediate layers such as recent Matformer~\cite{devvrit2024matformer}.} At test-time, Hi-Vec needs to compute multiple gradient norms for layer selection. However, to adapt to target data, it doesn't introduce any additional backward passes or parameters needed for backpropagation. Specifically, Hi-Vec uses only one of the hierarchical linear layers for backpropagation and during inference, ensuring no additional parameters are introduced. In Table~\ref{tab:inf_time}, we compare the inference times for common test-time adaptation methods with and without Hi-Vec integration based on a {single Nvidia A100 GPU}. Hi-Vec introduces a modest increase in compute time for Tent and SAR, with Tent + Hi-Vec taking 1m 25s compared to 31s for Tent and SAR + Hi-Vec taking 1m 42s compared to 50s for SAR. For STAMP, Hi-Vec results in a negligible change. Overall, Hi-Vec achieves robust adaptation with minimal computational overhead, making it useful for test-time adaptation.

\begin{table}[ht]
\begin{minipage}{0.26\textwidth}
\resizebox{\columnwidth}{!}{
\begin{tabular}{lr}
\toprule
\textbf{Methods} & \textbf{Time} \\ 
\midrule
Tent~\cite{wang2021tent} & 31s\\
 \rowcolor{lightblue}
Tent + \textbf{Hi-Vec}  & 1m 25s \\
SAR~\cite{niu2022towards} & 50s  \\
\rowcolor{lightblue}
SAR + \textbf{Hi-Vec} &  1m 42s \\

STAMP~\cite{yu2024stamp}
 & 11m 48s \\
\rowcolor{lightblue}
STAMP + \textbf{Hi-Vec}  &  11m 49s \\

\bottomrule
\end{tabular}}
\end{minipage} ~
\begin{minipage}{0.18\textwidth}
\caption{\textbf{The total inference time across all the fifteen domains} on Cifar-10-C with ResNet-18. The usage of Hi-Vec leads to a moderate increase in commute time for SAR and Tent. For STAMP, Hi-Vec introduces a negligible change in the inference time. }
\label{tab:inf_time}
\end{minipage}
\end{table}

\subsection{Mitigates Catastrophic Forgetting}
\label{app:continual}
\begin{figure}[t]
\centering
\includegraphics[width=0.97\linewidth, height=0.97\textheight, keepaspectratio]{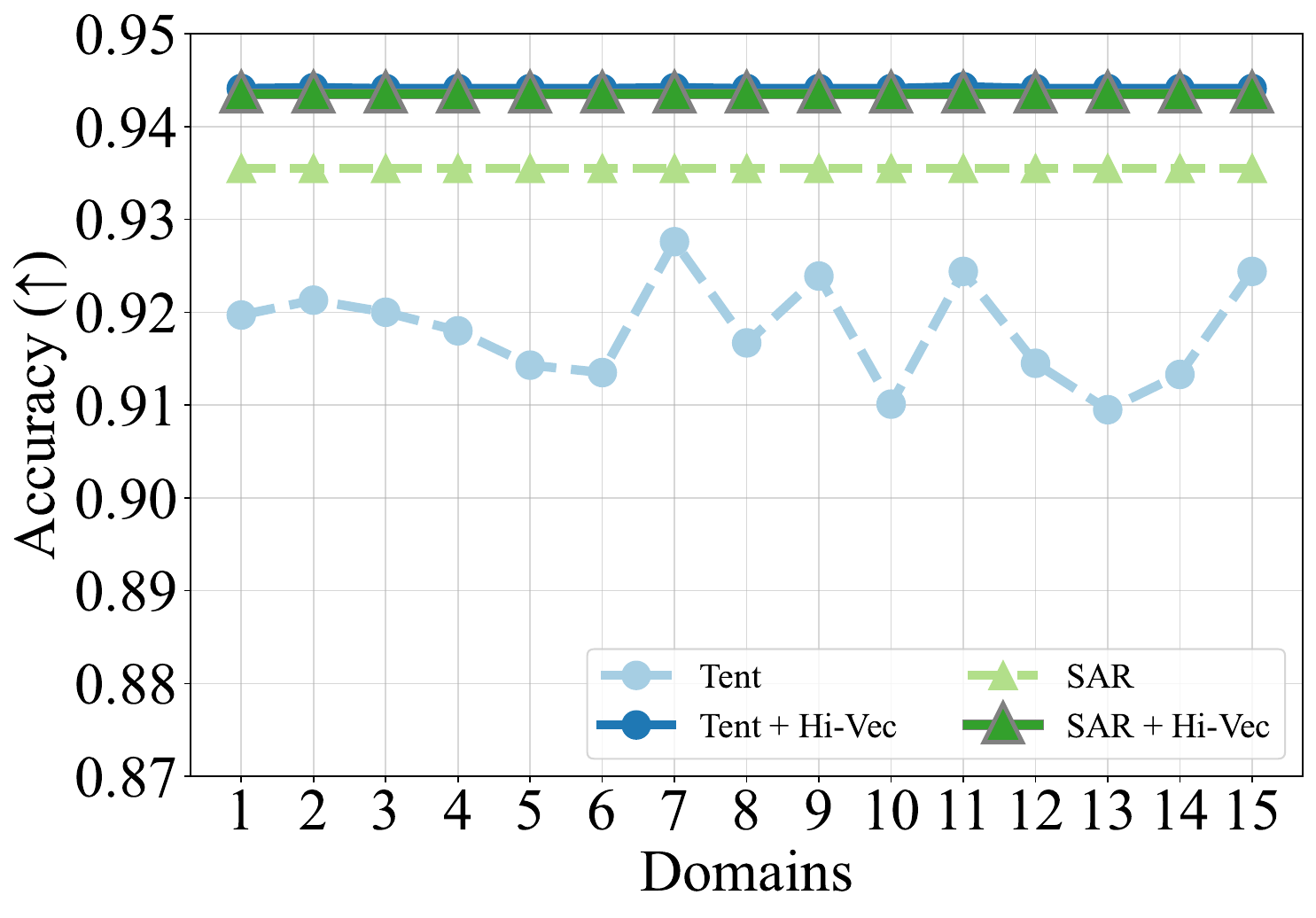}

\caption{\textbf{Mitigates Catastrophic Forgetting.} Resluts reported for Cifar-10-c using ResNet-18. We evaluate the model on the source domain after adapting it to every target domain. Hi-Vec preserves the source domain knowledge and prevents forgetting on the dataset at test-time. 
}

\label{fig:source_acc}
\vspace{-4mm}
\end{figure}

Retaining source domain knowledge during test-time adaptation is essential for leveraging the shared features between source and target domains~\cite{zhang2023adanpc}. Catastrophic forgetting, where the model loses previously acquired target information when adapting to new data, can severely degrade performance on source data samples. We also test Hi-Vec's capability for such cases and provide the results for catastrophic forgetting of source knowledge in  Fig~\ref{fig:source_acc}. In this setup, we adapt the model to a target domain and evaluate it on the source domain for the Cifar-10-c dataset with noise as an outlier with ResNet-18. Integrating Hi-Vec with methods such as SAR and Tent improves source domain accuracy during adaptation. Hi-Vec achieves this by employing its gradient-based layer selection mechanism to adapt only the most relevant hierarchical layer and its hierarchical layer agreement mechanism to prevent unnecessary updates for out-of-distribution samples. These Hi-Vec mechanisms preserve source knowledge while enabling robust adaptation to target shifts.

\subsection{Analysis of Hierarchical Layers Agreement and Benefits}
\label{app:mi_counts}
Hierarchical layers agreement, as detailed in the methodology and algorithm, enables Hi-Vec to skip adaptation for batches containing outliers, thereby preventing noisy model updates that could lead to model misspecification. This is particularly important because adapting to noisy OOD samples can destabilize the model by error accumulation and reinforcing irrelevant spurious features. Hierarchical layers in Hi-Vec capture coarse-to-fine representations, and their agreement reflects shared, meaningful features across these levels. When agreement is low, it indicates the absence of common features, signaling extreme OOD samples. To provide insights into this mechanism, Figure~\ref{fig:mi_count} shows the number of times adaptation was skipped during test-time using ResNet-18 on Cifar-10-c with noise as outlier data. The x-axis represents the domain count for the Cifar-10-c dataset with outliers during test-time adaptation, while the y-axis indicates the count of skipped adaptations. For Tent + Hi-Vec and STAMP + Hi-Vec, adaptation was skipped more frequently, focusing on inference instead, while SAR + Hi-Vec skipped adaptation for nearly half the batches. In comparison, common test-time adaptation~\cite{wang2021tent,yu2024stamp,lee2024entropy,liang2020we} do not skip adaptation (corresponding to 0 skips in the Figure~\ref{fig:mi_count}) and instead perform backpropagation for every batch. By avoiding adaptation on noisy batches that lead to error accumulation for the model and its predictions, Hi-Vec achieves stability on noisy target samples.

\subsection{Grad-CAM Visualizations and Layer selection insights }
\label{app:gradcam}

\begin{figure*}[t]
\centering
\includegraphics[width=1.0\linewidth, height=1.0\textheight, keepaspectratio]{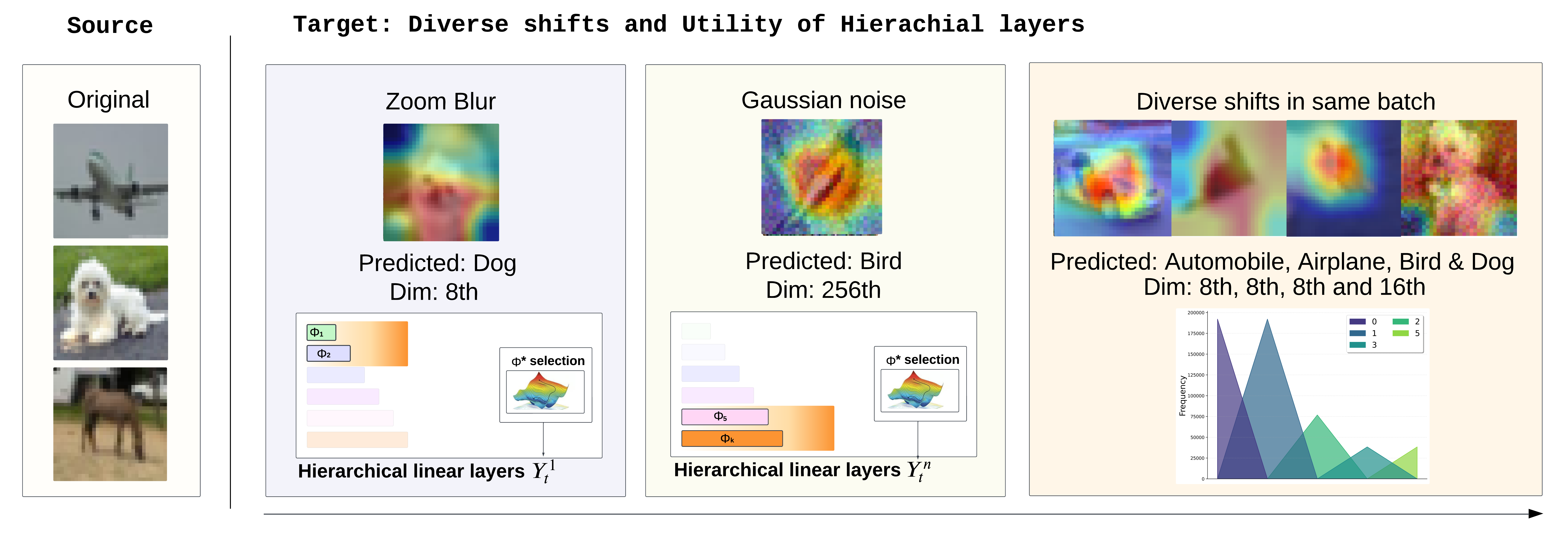}

\caption{\textbf{Grad-CAM Visualizations and Layer selection insights} on Cifar-10-c with ResNet-18 by Stamp + Hi-Vec. We provide the histogram figures for the outputs of the hierarchical linear layers. Together with the dimension of the model that is being used for the prediction and histogram of dimensions (where layer 0 has 8 dimensions, layer 1 has 16, and layer n has $2^{n+1}$ dimensions) for a random batch of the Cifar-10-c dataset. 
}

\label{fig:gradcam}
\vspace{-4mm}
\end{figure*}

\begin{figure}[h]
\centering
\includegraphics[width=0.97\linewidth, height=0.97\textheight, keepaspectratio]{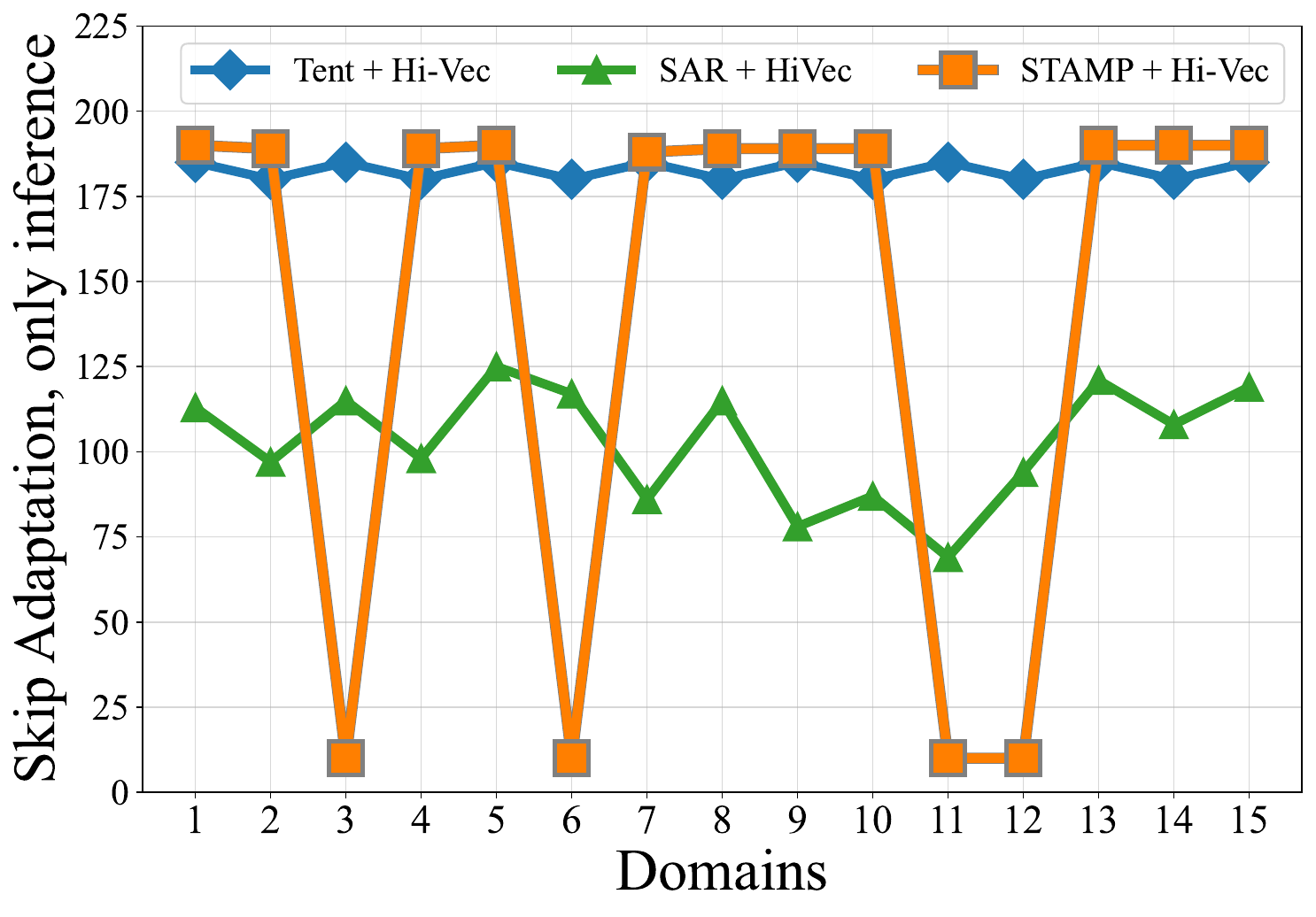}

\caption{\textbf{Analysis of Hierarchical Layers Agreement and Benefits.} Reported for the Cifar-10C dataset with a ResNet-18 encoder. We provide insights into how the hierarchical layers agreement works and how it counts. The common methods perform backpropagation on every batch (0 skips in the figure). Hi-Vec opts to skip adaptation due to the hierarchical layer agreement mechanism, improving the process by avoiding adaptation on noisy samples and noisy predictions. 
}

\label{fig:mi_count}
\vspace{-4mm}
\end{figure}

\begin{figure}[h]
\centering
\includegraphics[width=0.85\linewidth, height=0.9\textheight, keepaspectratio]{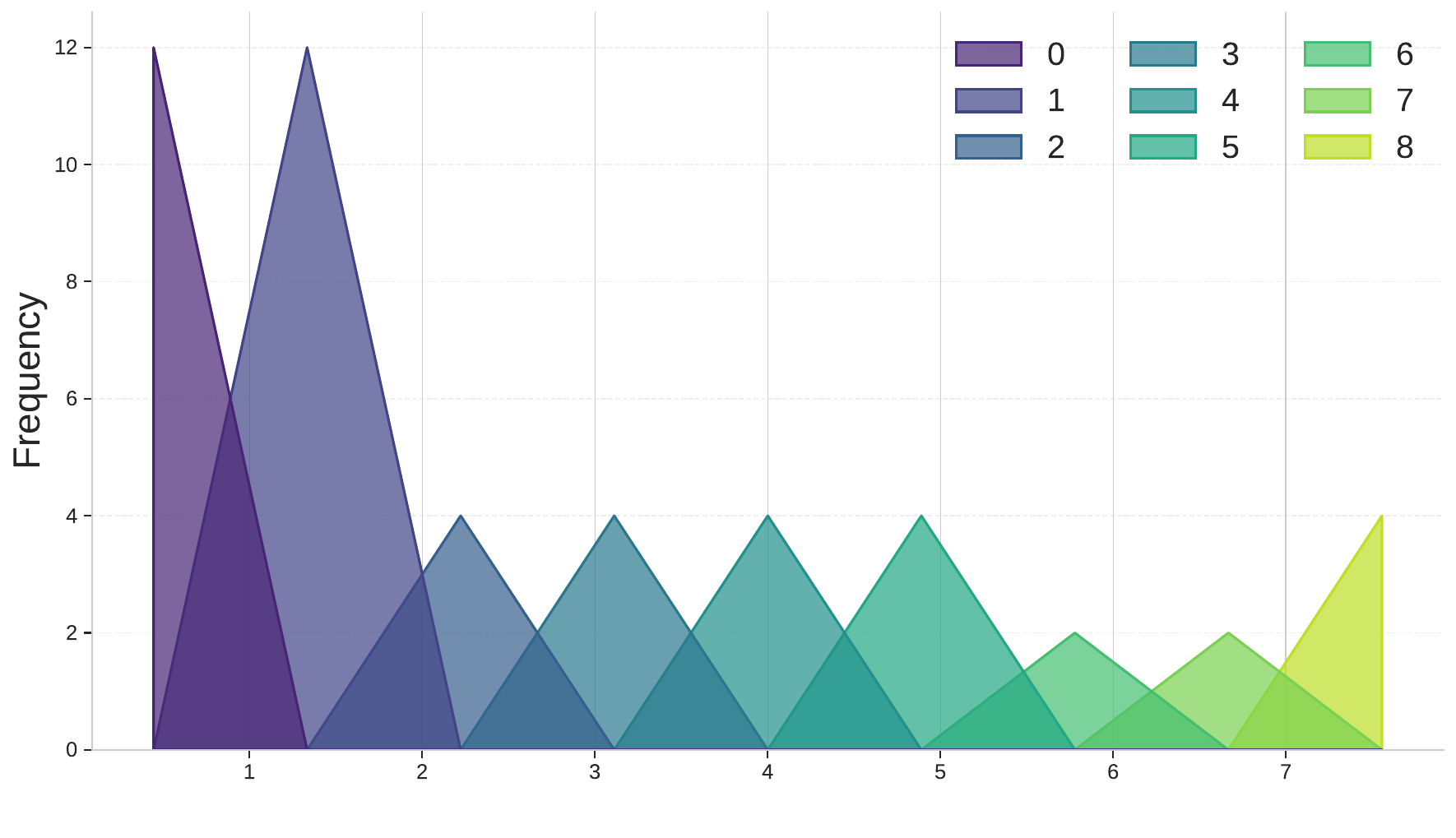}

\caption{\textbf{Layer selection insights.} We also provide the Layer selection insights with a histogram of counts for Waterbirds dataset by using DeYo + Hi-Vec with ResNet-18 at test-time. Layer 0 has 8 dimensions, layer 1 has 16, and layer n has $2^{n+1}$ dimensions. Hi-Vec dynamically selects layers across varying dimensions to address diverse distribution shifts effectively.
}

\label{fig:nmd_waterbirds}
\vspace{-4mm}
\end{figure}

At test-time, to analyze the behavior of hierarchical linear layers and the selection of \(\phi^{*}\) at test-time, we provide qualitative visualizations in Figure~\ref{fig:gradcam}. Specifically, we illustrate the Grad-CAM~\cite{selvaraju2020grad} outputs with STAMP + Hi-Vec for the selected hierarchical linear layer (\(\phi^{i}\)) alongside a histogram showing the frequency of selected layers across test batches of Cifar-10-c dataset with noise. Grad-CAM~\cite{selvaraju2020grad} visualizations highlight how different dimensions focus on distinct regions of the input image, enabling accurate classification by leveraging features relevant to the target distribution, even when it deviates significantly from the source. The histogram further demonstrates that multiple dimensions are utilized dynamically across test batches, reflecting Hi-Vec’s ability to adaptively select layers suited to varying shifts. This adaptive mechanism ensures that hierarchical representations effectively address diverse target shifts by focusing on features most relevant for each batch.
Additionally, In Figure~\ref{fig:nmd_waterbirds}, we also provide insights into layer selection for the Waterbirds dataset using DeYo + Hi-Vec for test-time adaptation. Unlike the shifts observed between CIFAR-10 and Cifar-10-c, the Waterbirds dataset introduces distinct distribution shifts that require different hierarchical layers to handle each batch effectively. The figure illustrates Hi-Vec's ability to dynamically select the optimal layer for each batch, enhancing the adaptability of test-time adaptation methods and ensuring robust performance across diverse shifts.

\section{Additional dataset and implementation details}

\subsection{Additional datasets information}
We train source models respectively on the source datasets as per the training and evaluation procedure as in ~\cite{wang2021tent,lee2024entropy,yu2024stamp,goyal2022test}. Akin to these common methods, the source datasets to train the respective source models include CIFAR-10, CIFAR-100, Waterbirds, and ImageNet as datasets. We utilize ResNet-18 and ResNet-50 with batch norm as the encoders. Our experimental setup involves multiple datasets at test time to evaluate the proposed method under two distinct types of diverse distribution shifts at test time. First, we consider outlier-aware scenarios where each test batch contains samples from Cifar-10-c~\cite{hendrycks2019benchmarking}, Cifar-100-c~\cite{hendrycks2019benchmarking}, or ImageNet-C~\cite{hendrycks2019benchmarking} datasets combined with a significant proportion of outlier datasets, including LSUN-C~\cite{yu2024stamp}, SVHN-C~\cite{yu2024stamp}, TinyImageNet-C~\cite{yu2024stamp}, Places-365-C~\cite{yu2024stamp}, and Textures-C~\cite{yu2024stamp}. For the in-distribution data, Cifar-10-c~\cite{hendrycks2019benchmarking} and Cifar-100-c~\cite{hendrycks2019benchmarking} include 15 corruption types (e.g., Gaussian noise, motion blur, fog) applied at five severity levels, with Cifar-10-c~\cite{hendrycks2019benchmarking} containing 10,000 images per corruption and Cifar-100-c~\cite{hendrycks2019benchmarking} containing 10,000 images per class. ImageNet-C~\cite{hendrycks2019benchmarking} serves as a large-scale benchmark with similar corruption types across 1,000 categories. We follow~\cite{yu2024stamp} to generate the outlier datasets. The outlier datasets are derived by applying the same 15 corruption types at the highest severity level of 5 to datasets such as LSUN, SVHN, TinyImageNet, Places-365, and Textures. In this setup, a single test batch contains both in-distribution samples from the corrupted CIFAR or ImageNet datasets and out-of-distribution samples from these corrupted outlier datasets. Next, we evaluate robustness to spurious correlations using datasets specifically designed to introduce misleading or non-causal correlations. The Waterbirds~\cite{sagawa2019distributionally} dataset consists of 11,788 images where spurious correlations arise due to the background (e.g., water or land) being associated with specific bird species. Similarly, ColoredMNIST~\cite{sagawa2019distributionally} modifies the MNIST dataset by introducing color as a spurious feature correlated with digit labels, consisting of 70,000 images across 10-digit classes. We train source models on Cifar-10, Cifar-100, Waterbirds, and ImageNet, respectively, using ResNet-18 and ResNet-50 as encoders. \purple{For ablations and additional experiments, we integrate Hi-Vec with re-implemented baselines from their official GitHub repositories.}

\subsection{Hyperparameters for the baselines}

We utilize the hyperparameters as provided in STAMP~\cite{yu2024stamp} and DeYo~\cite{lee2024entropy} repositories that include the implementation of other baseline methods such as Tent~\cite{wang2021tent}, SAR~\cite{niu2023towards}. As in these methods, we use ResNet-18 with batch norm for all our results. We list all the hyperparameters for the experiments that align with the original GitHub implementation repositories and implementations provided by STAMP\footnote{\url{https://github.com/yuyongcan/STAMP}}~\cite{yu2024stamp} and DeYo\footnote{\url{https://github.com/Jhyun17/DeYO/}}~\cite{lee2024entropy}.

\noindent \textbf{Tent.} ~\citet{wang2021tent} provided by~\cite{yu2024stamp} uses the Adam optimizer with beta set to 0.9 and with a momentum of 0.9.  For Cifar-10-c, the learning rate (\texttt{lr}) was set to 0.001. For Cifar-100-c, the learning rate was reduced to 0.0001. For ImageNet, the learning rate was further adjusted to 0.00025.

\noindent \textbf{SAR.} ~\citet{niu2023towards} provided by~\cite{yu2024stamp} uses the Adam optimizer with beta set to 0.9 and with a momentum of 0.9. For Cifar-10-c, the learning rate (\texttt{lr}) was set to 0.005 and the reset constant (\texttt{rst}) was 0.3. For Cifar-100-c, the learning rate was also 0.005, but \texttt{rst} was reduced to 0.2. For ImageNet, the learning rate was 0.0005 while \texttt{rst} was adjusted to 0.1. 

\noindent \textbf{STAMP.} ~\citet{yu2024stamp} uses the Adam optimizer with beta set to 0.9 with a momentum of 0.9.For Cifar-10-c, the learning rate (\texttt{lr}) was set to 0.1 and the stamp parameter (\texttt{alpha}) was 0.25. For Cifar-100-c, \texttt{lr} was set to 0.05 and \texttt{alpha} was 0.9. For ImageNet, \texttt{lr} was set to 0.01, and \texttt{alpha} was 0.8. The  MI threshold \(\tau_{\text{OOD}}\)\\ value of 1.6, Cosine threshold \(\tau\) of 0.6 was used for the datasets. We experimented with the hyperparameter values and have reported the values for which the performance is highest. 

\noindent \textbf{DeYo.} \citet{lee2024entropy} uses SGD as an optimizer with a momentum of 0.9. For the Waterbirds dataset, a pseudo-label threshold (\texttt{plpd\_threshold}) of 0.5, a deyo margin (\texttt{deyo\_margin}) of 0.5, an early deyo margin (\texttt{deyo\_margin\_e0}) of 0.4, a learning rate multiplier (\texttt{lr\_mul}) of 5, and an evaluation interval of 10 was applied. For ColoredMNIST, after pretraining a pseudo-label threshold of 0.5 was used, both deyo margins were set to 1.0,  and a learning rate multiplier of 5 and an evaluation interval of 30 has been utilized.

\noindent \textbf{Hi-Vec.} Our approach employs the hyperparameters listed above for the corresponding baselines for integration. Settings specific to our method involve the selection of a scaling parameter and a mutual information threshold classifying hierarchical layer agreement. For Cifar10, Imagenet, and Cifar100 experiments, we utilize a scaling ($\alpha$) parameter of 0.7 and a mutual information threshold \(\tau_{\text{OOD}}\) of 1.2 for the experiments. We have detailed the implementation of our method in Section 3, the detailed algorithm, and Figures 1 and 2 from the main paper. We will release the full code in the final version.  

\subsection{Results for the main paper with standard deviations. }
We also provide the standard deviations for the main results from the main paper.

\begin{table*}[t]
\centering
\small

\resizebox{0.9\textwidth}{!}{
\begin{tabular}{llcccccccccccc}
\toprule
                   &  & \multicolumn{3}{c}{\textbf{Noise}} & \multicolumn{3}{c}{\textbf{SVHN-C}} & \multicolumn{3}{c}{\textbf{LSUN-C}} & \multicolumn{3}{c}{\textbf{TinyImageNet-C}} \\ 
                   \cmidrule(r){3-5} \cmidrule(r){6-8} \cmidrule(r){9-11} \cmidrule(r){12-14}
                   &  \textbf{Methods} &  Acc     & AUC      &H-score      &       Acc     & AUC      &H-score      &        Acc     & AUC      &H-score      &     ACC     & AUC      &H-score      \\ \midrule
\multirow{10}{*}{\rotatebox{90}{\textbf{Cifar-10-C}}} & Source  &  57.3     & 70.4      &  62.3    &  57.3     &   67.4    &  61.1    &  57.3     &  62.8     &  59.6   &57.3       & 64.5     &59.4     \\
                    &  \bluet{MRL~\cite{kusupati2022matryoshka}}  &  59.7    &  65.7    & 62.6     &  59.7    &   64.4   & 62.0  &  59.7 &   61.9   & 60.2   & 59.7      & 64.1    & 61.8    \\

                   & BN Stats~\cite{nado2020evaluating}  &  72.9     & 68.6      & 70.6     &  78.7     &  75.3     & 76.9     & 79.4      & 79.4      &  79.4    &  79.0     & 72.9      & 75.8     \\

                   & EATA~\cite{niu2022efficient} &   72.9    &  68.5     &70.6      &78.8       & 75.3      & 76.9     &  79.4     &  79.4     &  79.4    &   78.9    &   73.1    &  75.9    \\

                   & CoTTA~\cite{wang2022continual} &  77.3     &  62.4     &  67.3    &    81.6   &    78.6   &  80.1    & 82.2      &     84.2  &  83.2    &   81.9    &   \textbf{75.3}    &  78.4    \\

                   & RoTTA~\cite{yuan2023robust}    &   77.6    &  74.3     &   75.6   &    78.4   &   76.0    &  77.2    &   78.8    &    79.5   & 79.1     &    78.6   &  73.3     &    75.8  \\

                   & SoTTA~\cite{gong2023sotta} &  77.8     &51.7       &61.6      &79.3       &72.8       &75.9      &79.8       & 77.9      & 78.8     & 79.6      & 72.6      & 75.9     \\

                   & OWTTT~\cite{li2023robustness} &  62.3     &  64.4     &  58.5    &    66.1   &   75.3    &  69.6    &    63.1   &        78.9    &   68.5    &    56.3   & 58.8     &  56.2     \\
                   \cmidrule(lr){2-14}
                 \cellcolor{white}
                   & Tent~\cite{wang2021tent} &  77.4     &  48.7     &  59.7    &     80.8  &   54.9    &   65.1   &    81.2   &     62.3  &   70.4   &   81.1    &  65.6     &    72.4  \\

                   & SAR~\cite{niu2022towards} &    72.9   &  68.5      &70.6      & 78.7       & 75.3      & 76.9     &  79.4     &  79.4    &79.4 & 79.0     & 72.9      &   75.8    \\

                   & STAMP~~\cite{yu2024stamp}&   {77.9}    &    {83.2}   &   {80.1}   &     {82.3}  &   {79.2}    &  {80.6}    &   {83.5}    &    {86.3}   &  {84.8}    &   {82.6}    &  74.9     &   {78.5}   \\ 

                    \cmidrule(lr){2-14}

                   \rowcolor{lightblue}
    &  \textit{Tent + {\textbf{Hi-Vec}}} &  
      \inc{80.7 \scriptsize{$\pm$0.2}} &  
      \inc{63.0 \scriptsize{$\pm$0.2}} &  
      \inc{70.5 \scriptsize{$\pm$0.2}} &  
      \inc{81.7 \scriptsize{$\pm$0.1}} &  
      \inc{55.4 \scriptsize{$\pm$0.1}} &  
      \inc{66.0 \scriptsize{$\pm$0.1}} &  
      \inc{81.7 \scriptsize{$\pm$0.2}} &  
      \inc{62.8 \scriptsize{$\pm$0.2}} &  
      \inc{71.0 \scriptsize{$\pm$0.2}} &  
      \inc{82.5 \scriptsize{$\pm$0.1}} &  
      \inc{65.8 \scriptsize{$\pm$0.1}} &  
      \inc{73.2 \scriptsize{$\pm$0.1}} \\

\rowcolor{lightblue}
    &  \textit{SAR + {\textbf{Hi-Vec}}} &  
      \inc{77.7 \scriptsize{$\pm$0.1}} &  
      \inc{63.8 \scriptsize{$\pm$0.1}} &  
      \inc{70.7 \scriptsize{$\pm$0.1}} &  
      \inc{82.5 \scriptsize{$\pm$0.1}} &  
      \inc{73.3 \scriptsize{$\pm$0.1}} &  
      \inc{77.7 \scriptsize{$\pm$0.1}} &  
      \inc{80.2 \scriptsize{$\pm$0.2}} &  
      \inc{79.9 \scriptsize{$\pm$0.2}} &  
      \inc{80.0 \scriptsize{$\pm$0.2}} &  
      \inc{83.0 \scriptsize{$\pm$0.1}} &  
      \inc{69.7 \scriptsize{$\pm$0.1}} &  
      \inc{75.7 \scriptsize{$\pm$0.1}} \\

\rowcolor{lightblue}
    &  \textit{STAMP + {\textbf{Hi-Vec}}} &  
      \inc{\textbf{83.6 \scriptsize{$\pm$0.1}}} &  
      \inc{\textbf{91.4 \scriptsize{$\pm$0.1}}} &  
      \inc{\textbf{87.3 \scriptsize{$\pm$0.1}}} &  
      \inc{\textbf{85.7 \scriptsize{$\pm$0.1}}} &  
      \inc{\textbf{82.7 \scriptsize{$\pm$0.1}}} &  
      \inc{\textbf{84.2 \scriptsize{$\pm$0.1}}} &  
      \inc{\textbf{84.3 \scriptsize{$\pm$0.2}}} &  
      \inc{\textbf{86.9 \scriptsize{$\pm$0.2}}} &  
      \inc{\textbf{85.5 \scriptsize{$\pm$0.2}}} &  
      \inc{\textbf{86.5 \scriptsize{$\pm$0.1}}} &  
      \inc{\textbf{81.1 \scriptsize{$\pm$0.1}}} &  
      \inc{\textbf{83.7 \scriptsize{$\pm$0.1}}} \\

                   \midrule

\multirow{10}{*}{\rotatebox{90}{\textbf{Cifar-100-C}}} & Source  &  35.8     &   43.1    &  38.0    &    35.8   &  49.4     &  40.1    &  35.8     &    58.2   &   43.2   &    35.8   &  57.1     & 42.7     \\
&  \bluet{MRL~\cite{kusupati2022matryoshka}} &  41.4    & 60.3     &  47.8   &  41.4     &   61.0   & 47.3  &  41.4 & 58.0     &  48.3   & 41.4      & 63.3    & 48.4    \\

                   & BN Stats~\cite{nado2020evaluating} &  45.8     &    80.9   &   58.4   &  52.7     &   72.5    &   60.9   &  53.7     &   73.8    &  62.0    &    53.2   &   68.6    &   59.7   \\

                   & EATA~\cite{niu2022efficient}  &    55.2   &  86.1     &   67.1   &     58.1  &   75.6    &  65.6    &  58.8    &    77.2   &   66.7   &    58.6   &   70.7    &    64.0  \\

                   & CoTTA~\cite{wang2022continual} &   47.0    & 83.4      &  59.9    &     53.7  &   73.2    &    61.8  &  54.3     &    76.9   &  63.6    &  54.5     &   68.1    &    60.4  \\

                   & RoTTA~\cite{yuan2023robust} &   47.9    &   54.0    &  49.4    &     47.3  &   67.0    &   55.3   &  48.3     &     69.5  &   56.7   &    47.8   &    65.5   &    55.0  \\

                   & SoTTA~\cite{gong2023sotta} &   54.4    &  53.3     & 52.8     &     53.6  &    70.3   &  60.7    &   54.4    &     70.8  &   61.4   &     53.9  &    68.4   &   60.1   \\

                   & OWTTT~\cite{li2023robustness} &  47.1     &  70.3     &   56.2   &      53.9 &    74.3   &   62.3   &   54.5    &      73.5 & 62.5     &   54.2    &    68.5   &   60.4   \\
                   \cmidrule(lr){2-14}
                   \cellcolor{white}  & Tent~\cite{wang2021tent}&   47.9    &   55.8    &   51.2   &  54.4     &  70.4     &  61.2    &    55.4   &    72.4   & 62.7     &   55.0    &    68.6   & 60.9     \\

                   & SAR~\cite{niu2022towards} &   57.5    &    88.6   &  68.9    &     59.2  &  65.2     &   61.9   &   60.5    &    73.5   &   66.3   &     60.8  &   72.1    & 65.9      \\

                   & STAMP~~\cite{yu2024stamp}&   {57.9}    & {98.4}      & {72.8}     &     {63.7}  &   {82.1}    &   {71.7}   &  {63.7}     &     \textbf{82.6}  &  {71.9}    &   {63.9}    &   {75.5}    &    {69.2}  \\ 
                \cmidrule(lr){2-14}
                & \textit{Tent + \textbf{Hi-Vec}} & 
              \inc{54.9 \scriptsize{$\pm$0.2}} & 
              \inc{68.2 \scriptsize{$\pm$0.2}} & 
              \inc{60.1 \scriptsize{$\pm$0.2}} & 
              \inc{54.7 \scriptsize{$\pm$0.2}} & 
              \inc{73.9 \scriptsize{$\pm$0.2}} & 
              \inc{62.3 \scriptsize{$\pm$0.2}} & 
              \inc{57.1 \scriptsize{$\pm$0.2}} & 
              \inc{72.7 \scriptsize{$\pm$0.2}} & 
              \inc{63.9 \scriptsize{$\pm$0.2}} & 
              \inc{55.3 \scriptsize{$\pm$0.1}} & 
              \inc{69.9 \scriptsize{$\pm$0.1}} & 
              \inc{61.1 \scriptsize{$\pm$0.1}} \\

                \rowcolor{lightblue}  &  \textit{SAR + \textbf{Hi-Vec}} & 
  \inc{57.9 \scriptsize{$\pm$0.1}} & 
  \inc{89.2 \scriptsize{$\pm$0.1}} & 
  \inc{69.4 \scriptsize{$\pm$0.1}} & 
  \inc{54.9 \scriptsize{$\pm$0.1}} & 
  \inc{73.4 \scriptsize{$\pm$0.1}} & 
  \inc{62.8 \scriptsize{$\pm$0.1}} & 
  \inc{60.9 \scriptsize{$\pm$0.2}} & 
  \inc{73.9 \scriptsize{$\pm$0.2}} & 
  \inc{66.7 \scriptsize{$\pm$0.2}} & 
  \inc{62.1 \scriptsize{$\pm$0.1}} & 
  \inc{74.4 \scriptsize{$\pm$0.1}} & 
  \inc{68.4 \scriptsize{$\pm$0.1}} \\

\rowcolor{lightblue} &  \textit{STAMP + \textbf{Hi-Vec}} & 
  \inc{\textbf{58.2 \scriptsize{$\pm$0.1}}} & 
  \inc{\textbf{89.6 \scriptsize{$\pm$0.1}}} & 
  \inc{\textbf{73.5 \scriptsize{$\pm$0.1}}} & 
  \inc{\textbf{64.4 \scriptsize{$\pm$0.1}}} & 
  \inc{\textbf{82.5 \scriptsize{$\pm$0.1}}} & 
  \inc{\textbf{72.1 \scriptsize{$\pm$0.1}}} & 
  \inc{\textbf{63.8 \scriptsize{$\pm$0.2}}} & 
  \inc{82.5 \scriptsize{$\pm$0.2}} & 
  \inc{\textbf{72.0 \scriptsize{$\pm$0.2}}} & 
  \inc{\textbf{64.6 \scriptsize{$\pm$0.1}}} & 
  \inc{\textbf{75.8 \scriptsize{$\pm$0.1}}} & 
  \inc{\textbf{70.4 \scriptsize{$\pm$0.1}}} \\

\bottomrule

\end{tabular}
}

\caption{\textbf{Results on adaptation with outlier datasets} using Cifar-10-C and Cifar-100-C as target datasets with four outlier datasets. We report the baselines and use evaluation metrics as provided by~\citet{yu2024stamp} with ResNet-18. Our results are averaged over fine runs.  Hi-Vec consistently improves performance (indicated by \incsymb{  } ) over the common methods and is the top-performer (\textbf{bold}) }
\label{table:cifar_tab1_std}
\end{table*}

\begin{table*}[t]
\vspace{-3mm}
\centering
\begin{minipage}[t]{0.31\textwidth}
\centering
\tiny %
\resizebox{\linewidth}{!}{%
\begin{tabular}{clcc}
\toprule
\textbf{Dataset} & \textbf{Methods}  & \textbf{Acc (\%)} & \textbf{Worst-Group Acc (\%)}  \\
\midrule
\multirow{9}{*}{\rotatebox{90}{\textbf{ColoredMNIST}}} 
    & Source                              & 63.40   & 20.05  \\
    & \bluet{MRL~\cite{kusupati2022matryoshka}}                              & 85.24  & 60.31 \\
    & Tent~\cite{wang2021tent}             & 57.06   & 9.80   \\
    & MEMO~\cite{zhang2021memo}             & 63.77   & 6.23   \\
    & SENTRY~\cite{prabhu2021sentry}         & 63.23   & 15.78  \\
    & EATA~\cite{niu2022efficient}          & 60.81   & 17.98  \\
    \cmidrule(lr){2-4}
    & SAR~\cite{niu2023towards}             & 58.37   & 12.36  \\
    & DeYO~\cite{lee2024entropy}            & 78.24   & 67.39  \\
    \cmidrule(lr){2-4}
    \rowcolor{lightblue}
    & \textit{SAR + \textbf{Hi-Vec}} & \inc{62.71 \tiny{$\pm$0.3}} & \inc{15.68 \tiny{$\pm$0.3}} \\
\rowcolor{lightblue}
& \textit{DeYO + \textbf{Hi-Vec}} & \inc{\textbf{79.53 \tiny{$\pm$0.2}}} & \inc{\textbf{68.62 \tiny{$\pm$0.2}}} \\

\midrule
\multirow{9}{*}{\rotatebox{90}{\textbf{WaterBirds}}}
    & Source                              & 83.16   & 64.90  \\
    & \bluet{MRL~\cite{kusupati2022matryoshka}}                              & 85.24  & 60.31 \\
    & Tent~\cite{wang2021tent}             & 82.95   & 54.14  \\
    & MEMO~\cite{zhang2021memo}             & 82.34   & 50.47  \\
    & SENTRY~\cite{prabhu2021sentry}         & 85.77   & 60.90  \\
    & EATA~\cite{niu2022efficient}          & 82.38   & 52.38  \\
    \cmidrule(lr){2-4}
    & SAR~\cite{niu2023towards}             & 82.60   & 53.41  \\
    & DeYO~\cite{lee2024entropy}            & 87.42   & 73.92  \\ 
    \cmidrule(lr){2-4}
    \rowcolor{lightblue}

    & \textit{SAR + {\textbf{Hi-Vec}}} & \inc{83.25 \tiny{$\pm$0.3}} & \inc{55.61 \tiny{$\pm$0.3}} \\ 
\rowcolor{lightblue}
& \textit{DeYO + \textbf{Hi-Vec}} & \inc{\textbf{89.53 \tiny{$\pm$0.2}}} & \inc{\textbf{77.23 \tiny{$\pm$0.2}}} \\

\bottomrule
\end{tabular}%
}
\caption{\textbf{Results for spurious correlation datasets} using ColoredMNIST and WaterBirds with ResNet-18. We report baselines provided by~\cite{lee2024entropy}. Our results are averaged over five runs. Hi-Vec improves (\incsymb{  } ) the common methods and performs the best ({bold}).   }
\label{tab:waterbirds_new}
\end{minipage}
\hfill
\begin{minipage}[t]{0.59\textwidth}
\centering
\tiny
\resizebox{\linewidth}{!}{%
\begin{tabular}{lcccccc}
\toprule
 & \multicolumn{3}{c}{\textbf{Places365-C}} & \multicolumn{3}{c}{\textbf{Textures-C}} \\ 
\cmidrule(r){2-4} \cmidrule(r){5-7}
\textbf{Methods} & ACC & AUC & H-score & ACC & AUC & H-score \\ \midrule
Source    & 18.2  & 61.6  & 26.1  & 18.2  & 54.6 & 25.8 \\
\bluet{MRL~\cite{kusupati2022matryoshka}}    & 18.4 & 61.9  & 28.3  & 18.4  & 54.6 & 27.4 \\
BN Stats~\cite{nado2020evaluating} & 31.1  & 67.7  & 41.1  & 31.6  & 61.2 & 40.7 \\
EATA~\cite{niu2022efficient} & {46.4}  & 72.6  & 56.0  & 46.4  & 62.2 & 52.8 \\
CoTTA~\cite{wang2022continual} & 33.8  & 66.9  & 43.5  & 34.2  & 60.7 & 42.8 \\
RoTTA~\cite{yuan2023robust} & 36.6  & 68.6  & 46.5  & 37.0  & 65.3 & 46.5 \\
SoTTA~\cite{gong2023sotta} & 41.7  & 67.8  & 50.7  & 41.8  & 60.3 & 48.8 \\
OWTTT~\cite{li2023robustness} & 9.1  & 54.0  & 13.9  & 9.4   & 59.4 & 14.6 \\
\midrule
Tent~\cite{wang2021tent} & 34.9  & 51.8  & 39.5  & 39.0  & 48.6 & 42.0 \\
SAR~\cite{niu2022towards} & 44.9  & 73.3  & 55.0  & 45.6  & 67.0 & 54.0 \\
STAMP & 46.4  & 77.7  & 57.6  & 46.5  & 71.9 & 56.2 \\
\midrule
\rowcolor{lightblue}

\textit{Tent + \textbf{Hi-Vec}} & 
  \inc{35.4 $\pm$0.4} & 
  \inc{52.0 $\pm$0.4} & 
  \inc{42.1 $\pm$0.4} & 
  \inc{39.4 $\pm$0.3} & 
  \inc{59.1 $\pm$0.3} & 
  \inc{47.2 $\pm$0.3} \\

\rowcolor{lightblue}
\textit{SAR + \textbf{Hi-Vec}}  & 
  \inc{45.3 $\pm$0.2} & 
  \inc{73.8 $\pm$0.2} & 
  \inc{56.1 $\pm$0.2} & 
  \inc{46.2 $\pm$0.3} & 
  \inc{67.4 $\pm$0.3} & 
  \inc{54.8 $\pm$0.3} \\

\rowcolor{lightblue}
\textit{STAMP + \textbf{Hi-Vec}} & 
  \inc{\textbf{46.9 $\pm$0.3}} & 
  \inc{\textbf{77.9 $\pm$0.3}} & 
  \inc{\textbf{58.5 $\pm$0.3}} & 
  \inc{\textbf{46.8 $\pm$0.2}} & 
  \inc{\textbf{71.7 $\pm$0.2}} & 
  \inc{\textbf{56.6 $\pm$0.2}} \\

\bottomrule
\end{tabular}%
}
\caption{\textbf{Results on adaptation with outlier datasets} using Imagenet-C with Places365-C and Textures-C as outlier datasets and ResNet-50. We report baselines and use evaluation metrics as provided by~\cite{yu2024stamp}. Our results are averaged over five runs. The conclusion is similar, improvement (\incsymb{  } ) over common methods and performs the best (bold). }
\label{table:imagenet_new_std}
\end{minipage}
\end{table*}

\section{Additional Related work }

\subsection{Test-time adaptation.} In addition to the aforementioned applications, test-time adaptation has also been used in vision language models~\cite{Bahri_2025_WACV,Zhang_2025_WACV,Sui_2025_WACV,A_Vargas_Hakim_2025_WACV}, continual learning~\cite{Imam_2025_WACV,Ma_2025_WACV,Yang_2024_CVPR}, classification that consider practical scenarios~\cite{Gu_2025_WACV,Colussi_2025_WACV,Dastmalchi_2025_WACV,Knott_2025_WACV,vray2025reservoirtta,lei2025ttvd,du2024unitta}. Moreover, recently,~\cite{kim2025testtime} proposed a framework that uses linear mode connectivity for adapted models during inference, with standard ResNet models.

\clearpage
\newpage

{
    \small
    \bibliographystyle{ieeenat_fullname}
    \bibliography{main}
}

\end{document}